\documentclass[10pt,journal]{IEEEtran}

\usepackage[pdftex]{graphicx}
\DeclareGraphicsExtensions{.pdf,.jpeg,.png}

\usepackage{amsmath, amsfonts, amsbsy, amssymb,color}
\usepackage{amsmath,amssymb}
\usepackage{amsthm}
\usepackage{multirow, adjustbox}

\usepackage{multirow,setspace,verbatim,amsfonts,graphicx,amsmath,amsthm,amsbsy,amssymb,epsfig,url}

\usepackage{epsfig}
\usepackage{graphicx}
\usepackage{amsmath}
\usepackage{amssymb}
\usepackage{color}

\usepackage{hyperref}

\newcommand{\Rd}{{\mathbb R}}

\newcommand{\hank}{\mathbb{H}}

\newcommand{\Hc}{{\mathcal H}}
\newcommand{\Qc}{{\mathcal Q}}

\newcommand{\rank}{\textsc{rank}}

\begin{document}

\title{ Framing U-Net via Deep Convolutional Framelets:
Application to Sparse-view  CT}
\author{Yoseob Han 
             and~Jong~Chul~Ye$^{*}$,~\IEEEmembership{Senior Member,~IEEE}
\thanks{Authors are with the Department of Bio and Brain Engineering, Korea Advanced Institute of Science and Technology (KAIST), 
		Daejeon 34141, Republic of Korea (e-mail: \{hanyoseob,jong.ye\}@kaist.ac.kr).}
\thanks{Part of this work was presented in 2017 International Conference on Fully Three-Dimensional Image Reconstruction in Radiology and Nuclear Medicine.}
}

\maketitle

\begin{abstract}
X-ray computed tomography (CT) using sparse projection views is a recent approach to reduce the radiation dose. However, due to the insufficient  projection views, an analytic  reconstruction approach using the filtered back projection (FBP) produces severe streaking artifacts. Recently, deep learning approaches using large receptive field neural networks such as  U-Net have demonstrated impressive performance for sparse-view CT reconstruction. However, theoretical justification is still lacking. 
Inspired by the recent theory of  {\em deep convolutional framelets}, 
the main goal of this paper is, therefore, to  reveal the 
limitation of U-Net and propose new multi-resolution deep learning schemes.
In particular, we show that the alternative U-Net variants such as dual frame and  the tight frame U-Nets satisfy the so-called frame condition which make them better for effective recovery of high frequency edges in
sparse view-CT. 
 Using extensive experiments with  real patient data set, we demonstrate that the new
 network architectures
  provide  better reconstruction performance.

\end{abstract}

\section{Introduction}

In X-ray CT,
due to the potential risk of radiation exposure, 
the main research thrust is  to reduce the  radiation dose.  Among various approaches for  low-dose CT, sparse-view CT is a recent proposal that lowers the
radiation dose by reducing the number of projection views \cite{sidky2008image,pan2009commercial,bian2010evaluation,ramani2012splitting,lu2011few,kim2015sparse,szczykutowicz2010dual,abbas2013effects,lee2016moving}. 
While  the sparse view CT may not be  useful
for existing multi-detector CTs  (MDCT) due to the  fast and continuous acquisition of projection views,  there are many
interesting new applications of sparse-view CT such as spectral CT using alternating kVp switching \cite{kim2015sparse,szczykutowicz2010dual},  dynamic beam blocker \cite{abbas2013effects,lee2016moving}, etc.
Moreover, in C-arm CT or dental CT applications, the scan time is  limited primarily by the relative slow speed of the plat-panel detector, rather than
the mechanical gantry speed,  so sparse-view CT
gives an opportunity to reduce the scan time \cite{pan2009commercial,bian2010evaluation}.

However,   insufficient 
projection views in sparse-view CT produces severe streaking artifacts in FBP reconstruction. To address this,  
researchers have investigated
compressed sensing approaches   \cite{donoho2006compressed} that minimize  the total variation (TV) or other sparsity-inducing penalties under a data fidelity  term \cite{sidky2008image,pan2009commercial,bian2010evaluation,ramani2012splitting,lu2011few,kim2015sparse,szczykutowicz2010dual,abbas2013effects,lee2016moving}. 
These approaches are, however, computationally  expensive due to the repeated applications of projection and back-projection during iterative update steps.

Recently, deep learning approaches have achieved tremendous success in various fields, such as classification \cite{krizhevsky2012imagenet}, segmentation \cite{ronneberger2015u}, denoising \cite{zhang2016beyond}, super resolution \cite{kim2015accurate, shi2016real},  etc.
In CT applications,  Kang et al \cite{kang2016deep} provided the first systematic study of deep  convolutional
neural network (CNN) for low-dose CT and showed that a deep CNN using directional wavelets is more efficient in removing low-dose related CT noises. This work was followed by many novel extensions  for low-dose CT \cite{chen2017low,kang2017wavresnet,kang2017wavedcf,chen2017lowBOE,adler2017learned,chen2017learned,wurfl2016deep,yang2017ct,wang2016perspective,yang2017low,wolterink2017generative}. Unlike these low-dose artifacts  from reduced tube currents, the streaking artifacts originated from sparse projection views show globalized artifacts that are difficult to remove using conventional denoising CNNs \cite{chen2015learning, mao2016image, xie2012image}. To address this problem, Jin et al \cite{jin2016deep} and Han et al \cite{han2016deep} independently proposed residual learning networks using U-Net \cite{ronneberger2015u}.
Because the streaking artifacts  are globally distributed,
CNN architecture with  large receptive field  was shown essential  in these works \cite{jin2016deep,han2016deep},
and their empirical performance was significantly better than the existing approaches.

In spite of such intriguing performance improvement by deep learning approaches, the origin
of the success  for inverse problems was poorly understood.
To address this,   we recently proposed 
so-called {\em deep convolutional framelets}  as a powerful mathematical framework to understand
 deep learning approaches for inverse problems \cite{ye2017deep}.
In fact, the  convolution framelets  was originally proposed by Yin et al \cite{yin2017tale} to generalize the low-rank Hankel matrix approaches
 \cite{ye2016compressive,jin2015annihilating,jin2018sparselow,jin2016general}
by representing a signal  using a fixed non-local basis convolved with data-driven local basis (the meaning of non-local and local bases will become clear later in this paper).
The novelty of our deep convolutional framelets was the discovery that   encoder-decoder network structure emerges from the
Hankel matrix decomposition \cite{ye2017deep}.
In addition, by controlling the number of filter channels, the neural network is trained to learn the optimal local bases so that it gives the best low-rank shrinkage \cite{ye2017deep}.
This discovery demonstrates an important link between the deep learning and the compressed sensing approach \cite{donoho2006compressed} through a Hankel structure matrix decomposition \cite{ye2016compressive,jin2015annihilating,jin2018sparselow,jin2016general}.

One of the key ingredients for  the deep convolutional framelets is the so-called {\em frame condition} for the non-local basis  \cite{ye2017deep}.
However,  we found that the existing U-Net architecture does not satisfy the frame condition and it overly emphasises
 the low frequency component of the signal \cite{ye2017deep}.
In the context of sparse-view CT, this artifact is manifested as blurring artifacts in the reconstructed images.
To address this problem, this paper investigates 
 two types of  novel network architectures that satisfy the frame condition.
First, we propose a {\em dual frame} U-Net architecture, in which 
the required modification is a  simple but intuitive additional by-pass connection
in the low-resolution path to generate a residual signal.
However, the dual frame U-Net is not optimal due to its relative large noise amplification factor.
To address this,  a {\em tight frame} U-Net with orthogonal wavelet frame is also proposed.
In particular, the  tight frame U-Net with Haar wavelet basis can be implemented by 
adding additional high-frequency path 
to the existing U-Net structure.
Our numerical experiments confirm that the dual frame and tight frame U-Nets
exhibit better high frequency recovery than the standard U-Net in sparse-view CT applications. 

Our source code and  test data set are can be found  at {https://github.com/hanyoseob/framing-u-net}. 

\section{Mathematical Preliminaries}
\label{sec:review}

\subsection{Notations}
For a matrix $A$, $R(A)$ denotes the range space of $A$, and $P_{R(A)}$ denotes the projection to the range space
of $A$.
The identity matrix is referred to as $I$.
 For a given matrix $A$, the notation $A^\dag$ refers to the generalized inverse. 
 The superscript $^{\top}$ of $A^{\top}$ denotes the Hermitian transpose. 
If a matrix $\Psi \in \Rd^{pd\times q}$ is partitioned as $\Psi = \begin{bmatrix} \Psi_1^\top & \cdots & \Psi_p^\top \end{bmatrix}^\top$ with sub-matrix $\Psi_i \in \Rd^{d\times q}$,
then $\psi_j^i$ refers to the $j$-th column of $\Psi_i$.
A vector $\overline v \in \Rd^n$ is referred to the flipped version of a vector $v \in \Rd^n$, i.e. its indices are reversed.
Similarly, for a given  matrix $\Psi \in \Rd^{d\times q}$, the notation
$\overline \Psi \in \Rd^{d\times q}$ refers to a matrix composed of flipped vectors, i.e.
$\overline \Psi = \begin{bmatrix} \overline \psi_1 & \cdots & \overline \psi_q \end{bmatrix}.$
For a block structured matrix  $\Psi \in \Rd^{pd\times q}$,  
with a slight abuse of notation, we define $\overline \Psi$ as
\begin{eqnarray}
\overline\Psi = \begin{bmatrix} \overline\Psi_1 \\ \vdots \\\overline \Psi_p \end{bmatrix},\quad \mbox{where}\quad \overline\Psi_i
=\begin{bmatrix} \overline {\psi_1^i} & \cdots & \overline {\psi_q^i} \end{bmatrix}  \in \Rd^{d\times q} . 
\end{eqnarray}

\subsection{Frame}

A family of functions $\{\phi_k\}_{k\in \Gamma}$ in a Hilbert space $H$ is called a {\em frame} if it satisfies the following inequality \cite{duffin1952class}:
\begin{eqnarray}\label{eq:frame0}
\alpha \|f\|^2 \leq \sum_{k\in \Gamma} |\langle f, \phi_k \rangle |^2  \leq \beta\|f\|^2,\quad \forall f \in H,
\end{eqnarray}
where $\alpha,\beta>0$ are called the frame bounds.  If $\alpha=\beta$, then the frame is said to be {tight}.
A frame is associated with a {frame operator} $\Phi$ composed of $\phi_k$:
$\Phi = \begin{bmatrix} \cdots & \phi_{k-1} & \phi_k & \cdots \end{bmatrix}.$
Then, \eqref{eq:frame0} can be equivalently written by
\begin{eqnarray}
\alpha \|f\|^2 \leq  \|\Phi^\top f\|^2  \leq \beta\|f\|^2,\quad \forall f \in H,
\end{eqnarray}
and the frame bounds can be  represented by
\begin{eqnarray}
\alpha = \sigma_{\min} (\Phi\Phi^\top),\quad \beta = \sigma_{\max}(\Phi\Phi^\top),
\end{eqnarray}
where $\sigma_{\min}(A)$ and $\sigma_{\max}(A)$ denote the minimum and maximum singular values of $A$, respectively. 
When the frame lower bound $\alpha$ is  non-zero, then the recovery of the original signal can be done 
from the frame coefficient $c=\Phi^\top f$ using the
{dual frame} $\tilde \Phi$ satisfying the so-called {\em frame condition}:
\begin{eqnarray}\label{eq:framecond0}
\tilde\Phi \Phi^\top = I,
\end{eqnarray}
because we have
$\hat f = \tilde \Phi c = \tilde \Phi \Phi^\top f = f.$
The explicit form of the dual frame is given by the pseudo-inverse:
\begin{eqnarray}\label{eq:dual}
\tilde \Phi = (\Phi\Phi^\top)^{-1} \Phi.
\end{eqnarray}
If  the frame coefficients  are contaminated by the noise $w$, i.e. $c=\Phi^\top f+w$, then
the recovered signal using the dual frame is given by
$\hat f = \tilde\Phi c= \tilde\Phi (\Phi^\top f+w) =  f + \tilde \Phi w.$
Therefore, the {\em noise amplification factor} can be computed  by
\begin{eqnarray}
 \frac{\|\tilde \Phi w\|^2}{\|w\|^2} = \frac{\sigma_{\max}(\Phi\Phi^\top)}{\sigma_{\min}(\Phi\Phi^\top)} = \frac{\beta}{\alpha}  = \kappa(\Phi\Phi^\top),
 \end{eqnarray}
where $\kappa(\cdot)$ refers to the condition number.
A tight frame has the minimum noise amplification factor, i.e.  $\beta/\alpha=1$,  and it
 is equivalent to the condition:
\begin{eqnarray}\label{eq:tight}
\Phi^\top \Phi = cI, \quad c>0.
\end{eqnarray}

\subsection{Hankel Matrix}

Since the Hankel matrix is an essential component in the theory of deep convolutional framelets \cite{ye2017deep},
 we  briefly review it to make this paper self-contained.
Here,  to avoid special treatment of boundary condition, our theory is mainly derived using the circular convolution. For simplicity,
we consider 1-D signal processing, but the extension to 2-D is straightforward \cite{ye2017deep}.

Let $f=[f[1],\cdots, f[n]]^T\in \Rd^n$ be the signal vector. Then,
a wrap-around Hankel matrix $\hank_d(f)$ is defined by
 \begin{eqnarray} 
\hank_d(f) =\left[
        \begin{array}{cccc}
        f[1]  &   f[2] & \cdots   &   f[d]   \\
       f[2]  &   f[3] & \cdots &     f[d+1] \\
         \vdots    & \vdots     &  \ddots    & \vdots    \\
              f[n]  &   f[1] & \cdots &   f[d-1] \\
        \end{array}
    \right] , 
    \end{eqnarray}
where $d$ denotes the matrix pencil parameter.
For a given multi-channel signal
\begin{eqnarray}
F:= [f_1\cdots f_p] \in \Rd^{n\times p} \ ,
\end{eqnarray}
 an {extended Hankel matrix}  is constructed by stacking  Hankel matrices side by side: 
\begin{eqnarray}
\hank_{d|p}\left(F\right)  := \begin{bmatrix} \hank_d(f_1) & \hank_d(f_2) & \cdots & \hank_d(f_p) \end{bmatrix}  \  .
\end{eqnarray}
As explained in \cite{ye2017deep}, the
Hankel matrix is closely related to the convolution operations in CNN.
Specifically,  for  a given convolutional filter $\overline\psi=[\psi[d],\cdots, \psi[1]]^T\in\Rd^d$,
a single-input single-output  convolution in CNN  can be represented using a Hankel matrix:
\begin{eqnarray}\label{eq:SISO}
y = f\circledast \overline\psi &=&\hank_d(f) \psi \quad \in \Rd^n \ .
\end{eqnarray}
Similarly, a single-input multi-ouput  convolution using CNN filter kernel $\Psi =[\psi_1 \cdots, \psi_q] \in \Rd^{d\times q}$ can be represented by
\begin{eqnarray}\label{eq:SIMO}
Y = f \circledast \overline\Psi = 
\hank_d(f)  \Psi \quad \in \Rd^{n\times q},
\end{eqnarray}
where  
$q$ denotes the number of output channels.
A multi-input multi-output convolution in CNN is represented  by
\begin{eqnarray}\label{eq:multifilter}
Y &=& F \circledast \Psi 
= \hank_{d|p}\left(F\right) \begin{bmatrix} \Psi_1 \\ \vdots \\ \Psi_p \end{bmatrix},
\end{eqnarray}
where $p$ and $q$ refer to the number of input and output channels, respectively, and 
\begin{eqnarray}
 \Psi_j =  \begin{bmatrix} \psi_1^j  & \cdots & \psi_q^j \end{bmatrix} \in \Rd^{d\times q}
 \end{eqnarray}
denotes the $j$-th input channel filter.
The extension to  the multi-channel 
2-D convolution operation for an image domain CNN  is straight-forward,
since similar matrix vector operations can be also used.
Only required change is the definition of the (extended) Hankel matrices, which is  defined as
{\em block} Hankel matrix.  For a more detailed 2-D CNN convolution operation in the form of Hankel matrix,
see \cite{ye2017deep}.

One of the most intriguing properties of the Hankel matrix is that it often has a low-rank structure and its low-rankness is related to the sparsity in the Fourier domain \cite{ye2016compressive,jin2015annihilating,jin2018sparselow}.
This property is extremely useful, as evidenced by their applications for many inverse problems and low-level computer vision problems \cite{jin2015annihilating,jin2018sparselow,jin2016general,ongie2016off,lee2016acceleration,lee2016reference,jin2016mri}.
Thus, we claim that
this property is  one of the origins of the success of deep learning for  inverse problems \cite{ye2017deep}.

\subsection{Deep Convolutional Framelets: A Review}

To understand this claim, we briefly review the theory of deep convolutional framelets \cite{ye2017deep} to make this paper  self-contained.
Specifically, inspired by the existing Hankel matrix approaches  
 \cite{jin2015annihilating,jin2018sparselow,jin2016general,ongie2016off,lee2016acceleration,lee2016reference,jin2016mri},
we consider the following regression problem: 
\begin{eqnarray}
\min_{f\in \Rd^{n}}  & \|f^* -f\|^2 \notag\\
\mbox{subject to }  &\quad \rank \hank_d(f) = r < d . \label{eq:fcost}
\end{eqnarray}
where $f^*\in \Rd^d$ denotes the ground-truth signal and $r$ is the rank of the Hankel structured matrix.
The classical approach to address this problem is to use singular value shrinkage or
matrix factorization \cite{jin2015annihilating,jin2018sparselow,jin2016general,ongie2016off,lee2016acceleration,lee2016reference,jin2016mri}.
However, in deep convolutional framelets \cite{ye2017deep}, the problem is addresssed  using  learning-based signal representation.

More specifically,
 for any feasible solution $f$ for \eqref{eq:fcost}, its Hankel structured matrix $\hank_d(f)$
 has  the singular value decomposition
$\hank_{d}(f) = U \Sigma V^{\top}$
where $U =[u_1 \cdots u_r] \in \Rd^{n\times r}$ and $V=[v_1\cdots v_r]\in \Rd^{d\times r}$ denote the left and the right singular vector bases matrices, respectively;
$\Sigma=(\sigma_{ij})\in\mathbb{R}^{r\times r}$ is the diagonal matrix with singular values.  
Now, consider the  matrix pairs $\Phi, \tilde \Phi \in \Rd^{n\times n}$  satisfying the {\em frame condition}:
\begin{eqnarray}\label{eq:framecond}
\tilde \Phi \Phi^\top = I . 
\end{eqnarray}
These bases are refered to as {\em non-local bases} since they interacts with all  the $n$-elements of $f\in \Rd^n$  by multiplying them to the left 
of $\hank_d(f) \in \Rd^{n\times d}$ \cite{ye2017deep}.
In addition,  we need another matrix pair  $\Psi$, $\tilde \Psi\in \Rd^{d\times r}$ satisfying the low-dimensional subspace constraint:
\begin{eqnarray}\label{eq:projection}
 \Psi \tilde \Psi^{\top} = P_{R(V)} .
 \end{eqnarray}
These are called {\em local bases} because it only interacts with $d$-neighborhood of the signal $f \in \Rd^n$ \cite{ye2017deep}.
Using Eqs.~\eqref{eq:framecond} and \eqref{eq:projection},  we can obtain the following matrix equality:
\begin{eqnarray}
\hank_d(f) = \tilde \Phi \Phi^\top  \hank_d(f)  \Psi \tilde \Psi^{\top}. 
\end{eqnarray}
Factorizing $ \Phi^\top  \hank_d(f)  \Psi $ from the above equation results in 
 the decomposition of $f$ using a
single layer encoder-decoder architecture \cite{ye2017deep}:
\begin{eqnarray}
f = 
\left(\tilde\Phi C\right) \circledast \nu(\tilde \Psi),  &&
C = \Phi^\top \left( f \circledast \overline \Psi  \right)  \label{eq:finsuf},
\end{eqnarray}
where  the encoder and decoder convolution filters   are respectively given by
\begin{eqnarray}
\overline\Psi := \begin{bmatrix} \overline\psi_1 & \cdots & \overline\psi_q \end{bmatrix} \in \Rd^{d\times q} , ~
\nu(\tilde\Psi) :=   \frac{1}{d} \begin{bmatrix}  \tilde \psi_1 \\ \vdots  \\
\tilde \psi_q
\end{bmatrix}\in \Rd^{dq} . 
\end{eqnarray}

Note that 
\eqref{eq:finsuf} is the general form of the signals that are associated with a rank-$r$ Hankel structured matrix,
and we are interested in specifying bases for  optimal performance.
In the theory of deep convolutional framelets \cite{ye2017deep},  $\Phi$ and $\tilde \Phi$ correspond to the user-defined
generalized pooling and unpooling to satisfy the frame condition \eqref{eq:framecond}.
On the other hand, the filters $\Psi$, $\tilde \Psi$ need to be estimated from the data.
To limit the search space for the filters, 
we consider $\Hc_0$, which consists of signals  that have positive framelet coefficients: 
\begin{eqnarray}\label{eq:H0}
\Hc_0 &=& \left\{ f\in \Rd^n|f = \left(\tilde\Phi C\right) \circledast \nu(\tilde \Psi), \right.  \notag \\
 && \left. C = \Phi^\top \left( f \circledast \overline \Psi \right),  [C]_{kl} \geq 0,~\forall k,l \right\} \  ,
\end{eqnarray}
where $[C]_{kl}$ denotes the $(k,l)$-th element of the matrix $C$.
Then, the main goal of the neural network training  is to learn   ($\Psi$, $\tilde \Psi$) from training data  $\{(f_{(i)}, f_{(i)}^*)\}_{i=1}^N$ assuming that
$\{f_{(i)}^*\}$ are associated with rank-$r$ Hankel matrices.
More specifically,  our regression problem for the training data  under low-rank Hankel matrix constraint in \eqref{eq:fcost}  is given by
\begin{eqnarray}
\min_{\{f_{(i)}\}\in \Hc_0}   \sum_{i=1}^N\|f_{(i)}^* - f_{(i)}\|^2, 
\end{eqnarray}
which can be equivalently represented by
\begin{eqnarray}\label{eq:newcost}
  \min_{( \Psi, \tilde\Psi)}  \sum_{i=1}^N\left\|f_{(i)}^* - \Qc(f_{(i)};\Psi,\tilde\Psi)\right\|^2 ,
\end{eqnarray}
where
\begin{eqnarray}
\Qc(f_{(i)};\Psi,\tilde\Psi)= \left(\tilde\Phi C[f_{(i)}]\right) \circledast \nu(\tilde \Psi)\\
\quad C[f_{(i)}] =\rho\left(
\Phi^\top \left( f_{(i)} \circledast \overline \Psi  \right)\right),
\end{eqnarray}
where $\rho(\cdot)$ is the ReLU to impose the positivity for the framelet coefficients.
After the network is fully trained, the inference for a given noisy input $f$ is simply done by
$\Qc(f;\Psi,\tilde\Psi)$, which is equivalent to find a denoised solution that has the rank-$r$ Hankel structured matrix.

In the sparse-view CT problems, it was consistently shown that the residual learning with by-pass connection is better than
direct image learning \cite{jin2016deep,han2016deep}.
To investigate this phenomenon systematically,
assume that the input image $f_{(i)}$ from sparse-view CT is contaminated with streaking artifacts:
\begin{eqnarray}
f_{(i)} = f_{(i)}^*+h_{(i)}, 
\end{eqnarray}
where $h_{(i)}$ denotes the streaking artifacts and $f_{(i)}^*$ refers to the artifact-free ground-truth.
Then, instead of using the cost function \eqref{eq:newcost}, 
the residual network training \eqref{eq:newcost} is formulated as \cite{han2016deep}:
\begin{eqnarray}\label{eq:training2}
 \min_{( \Psi, \tilde\Psi)}  \sum_{i=1}^N\left\|h_{(i)}- \Qc(f_{(i)}^*+h_{(i)};\Psi,\tilde\Psi)\right\|^2 .
 \end{eqnarray}
In \cite{ye2017deep}, we showed that
this residual learning scheme is to find the filter
$\overline \Psi $ which approximately annihilates the true signal $f_{(i)}^*$, i.e. 
\begin{eqnarray}\label{eq:fanal}
 f_{(i)}^*\circledast \overline \Psi \simeq 0   \  , 
\end{eqnarray}
such that the signal decomposition using deep convolutional framelets can be applied for the streaking artifact signal, i.e,
\begin{eqnarray}
\left(\tilde\Phi C[f_{(i)}^*+h_{(i)}]\right) \circledast \nu(\tilde \Psi)  &\simeq& \left(\tilde\Phi C[h_{(i)}]\right) \circledast \nu(\tilde \Psi) \notag \\
&=& h_{(i)} \  .
\end{eqnarray}
Here,  the first approximation comes from  
\begin{eqnarray}
 C[f_{(i)}^*+h_{(i)}] = \Phi^\top \left( (f_{(i)}^*+h_{(i)})\circledast \overline \Psi  \right)  \simeq C[h_{(i)}]
 \end{eqnarray}
 thanks to the annihilating property \eqref{eq:fanal}.
Accordingly, the neural network is trained to
 learn the structure of the true image  to annihilate them, but still to retain the artifact signals.

The idea can be further extended to the multi-layer deep convolutional framelet expansion.
More specifically, for the $L$-layer decomposition,
 the space $\Hc_0$ in \eqref{eq:H0} is now recursively defined as:
\begin{eqnarray}
\Hc_0 &=& \left\{ f\in \Rd^n|f = \left(\tilde\Phi C\right) \circledast \nu(\tilde \Psi), \right. \notag \\
 && \left. C = \Phi^\top \left( f \circledast \overline \Psi \right),~[C]_{kl}\geq 0,\forall k,l,~C \in \Hc_1 \right\} \notag\\
 \end{eqnarray}
 where $\Hc_l,~ l=1,\cdots, L-1$ is defined as
 \begin{eqnarray}
 \Hc_l &=& \left\{ Z\in \Rd^{n\times p_{(l)}}|Z = \left(\tilde\Phi C^{(l)}\right) \circledast \nu(\tilde \Psi^{(l)}), \right.  \notag \\
 && \left. C^{(l)} = \Phi^\top \left(Z \circledast \overline \Psi^{(l)} \right),~[C]_{kl}\geq 0,\forall k,l,\right. \notag \\
 &&\left. ~C^{(l)} \in \Hc_{l+1} \right\}  \notag \\
 \Hc_L &=& \Rd^{n\times p_{(L)}} ,
\end{eqnarray}
where the $l$-th layer encoder and decoder  filters are now defined by
\begin{eqnarray}
\overline\Psi^{(l)} &:=&\begin{bmatrix}   \overline\psi_1^{1} & \cdots &   \overline\psi_q^{1}  \\ \vdots & \ddots & \vdots \\
\overline\psi_1^{p_{(l)}} & \cdots &  \overline\psi_{q_{(l)}}^{p_{(l)}} 
\end{bmatrix}  \in \Rd^{d_{(l)}p_{(l)} \times q_{(l)}} 
\end{eqnarray}
\begin{eqnarray}
\nu(\tilde\Psi^{(l)}) :=  \frac{1}{d} \begin{bmatrix}  \tilde \psi_1^1 & \cdots &  \tilde \psi_1^{p_{(l)}}  \\ \vdots & \ddots & \vdots \\
\tilde \psi_{q_{(l)}}^1 & \cdots &  \tilde\psi_{q_{(l)}}^{p_{(l)}} 
\end{bmatrix}  \in \Rd^{d_{(l)}q_{(l)} \times p_{(l)}}
\end{eqnarray}
and $d_{(l)}, p_{(l)}, q_{(l)}$ denote the filter length, and the number of input and output channels,  respectively.
By recursively narrowing the search space of the convolution frames in each layer as described above, we can obtain the deep convolution framelet extension and the associated training scheme. For more details, see \cite{ye2017deep}.

In short,  one of the most important observations in \cite{ye2017deep} is that the non-local bases $\Phi^\top$ and $\tilde\Phi$ correspond to the {generalized pooling
and unpooling} operations, while the local basis $\Psi$ and $\tilde \Psi$ work as learnable convolutional filters.
Moreover,  for the generalized pooling operation,  
the frame condition \eqref{eq:framecond} is the most important  prerequisite for enabling the recovery condition and controllable shrinkage behavior,  which is the main  criterion for constructing our U-Net variants in the next section.

\section{Main Contribution}
\label{sec:main}

\subsection{U-Net for Sparse-View CT and Its Limitations}

\begin{figure}[!bt]
    \centerline{\includegraphics[width=0.7\linewidth]{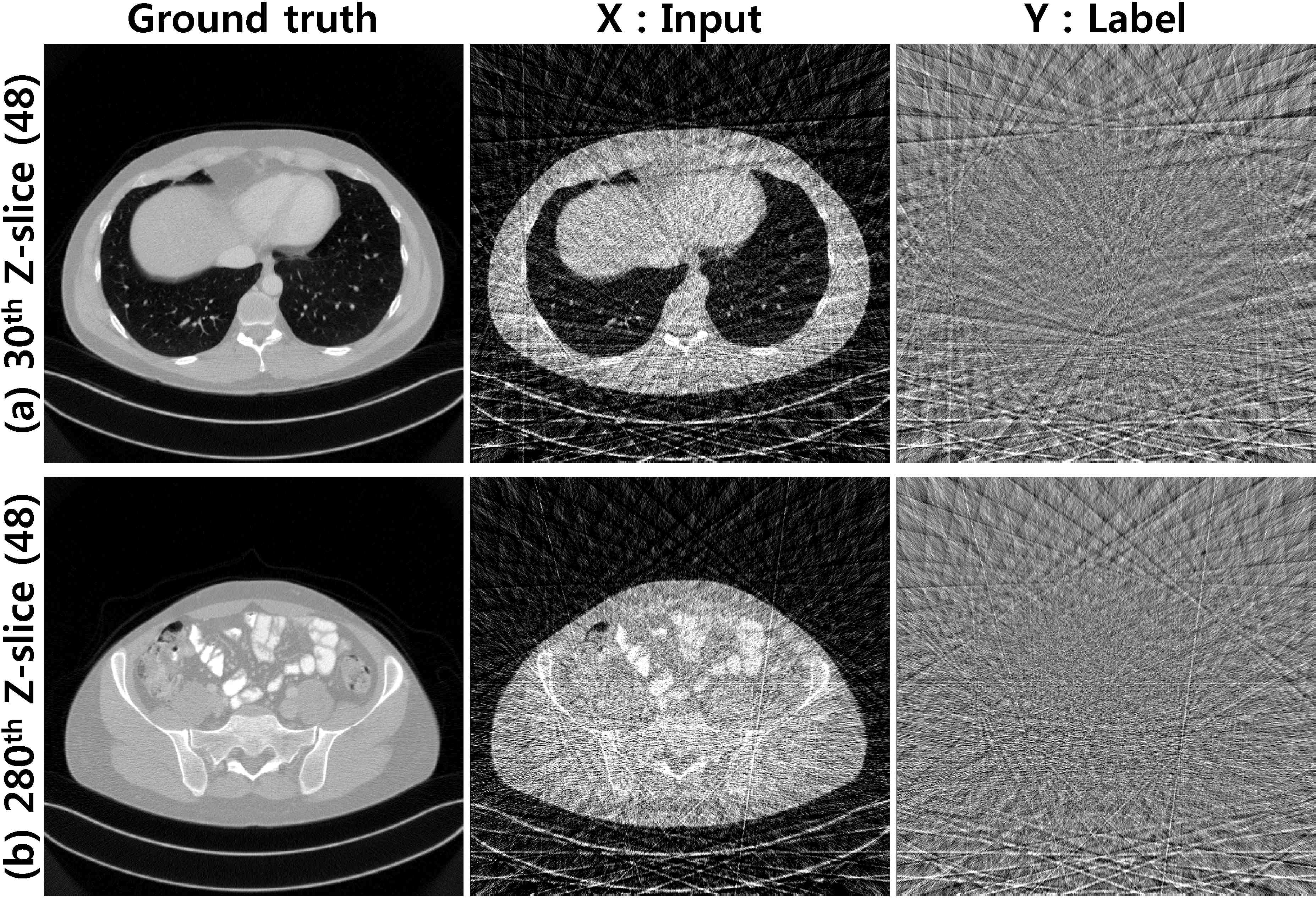}}
    \vspace*{-0.2cm}
    \caption{CT streaking artifact patterns in the reconstruction images  from 48 projection views.  }
    \label{fig:streaking_pattern}
\end{figure}

Figs. \ref{fig:streaking_pattern}(a)(b) show  two reconstruction images and their artifact-only images when only 48 projection views are available.
There is a significant streaking artifact that emanates from images over the entire image area.
This suggests that the receptive field of the convolution filter should cover the entire area of the image to effectively suppress the streaking artifacts.

   \begin{figure}[!b]
       \vspace*{-0.2cm}
    \centerline{\includegraphics[width=4.cm]{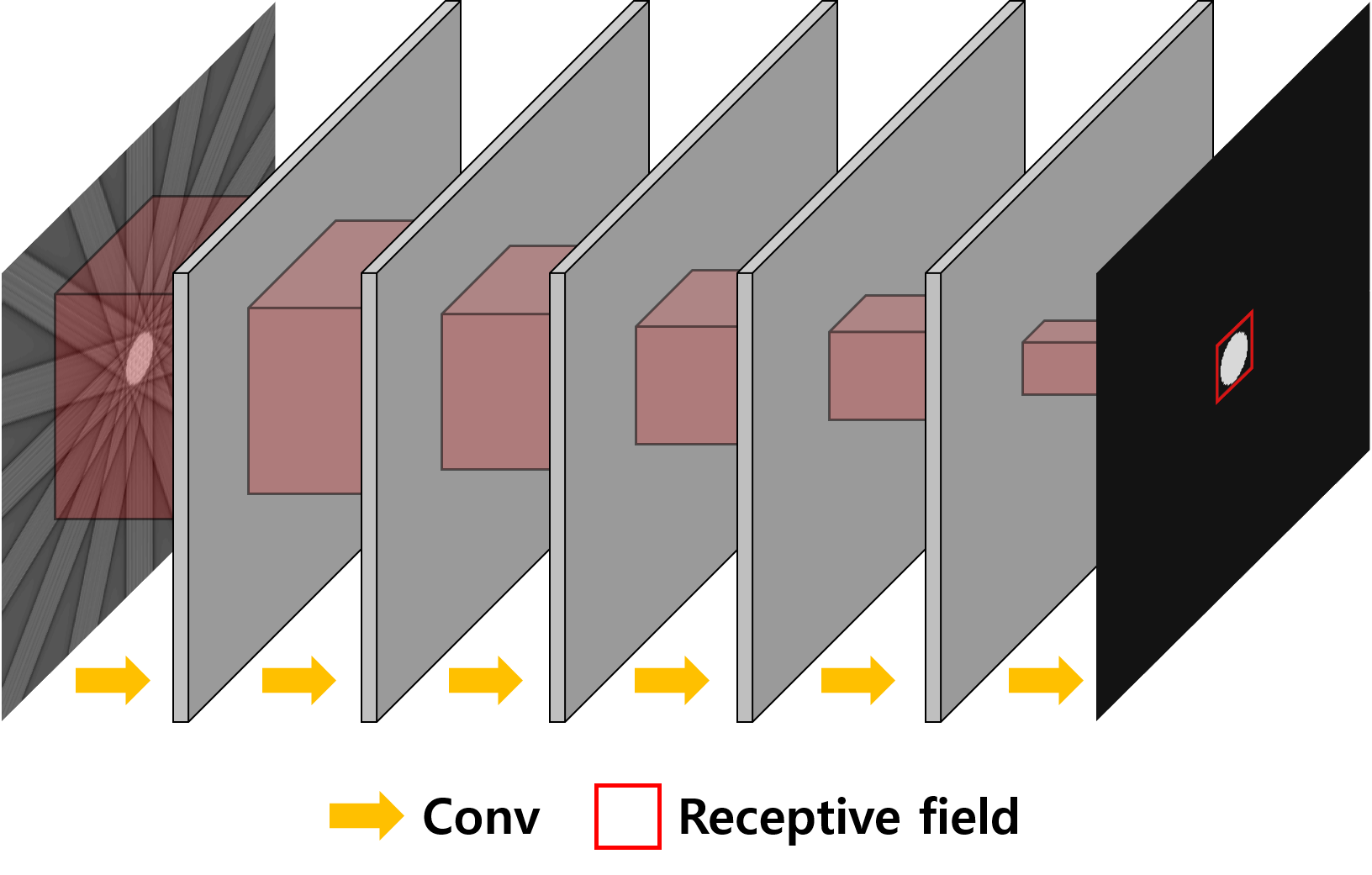}\hspace{0.5cm}\includegraphics[width=4.cm]{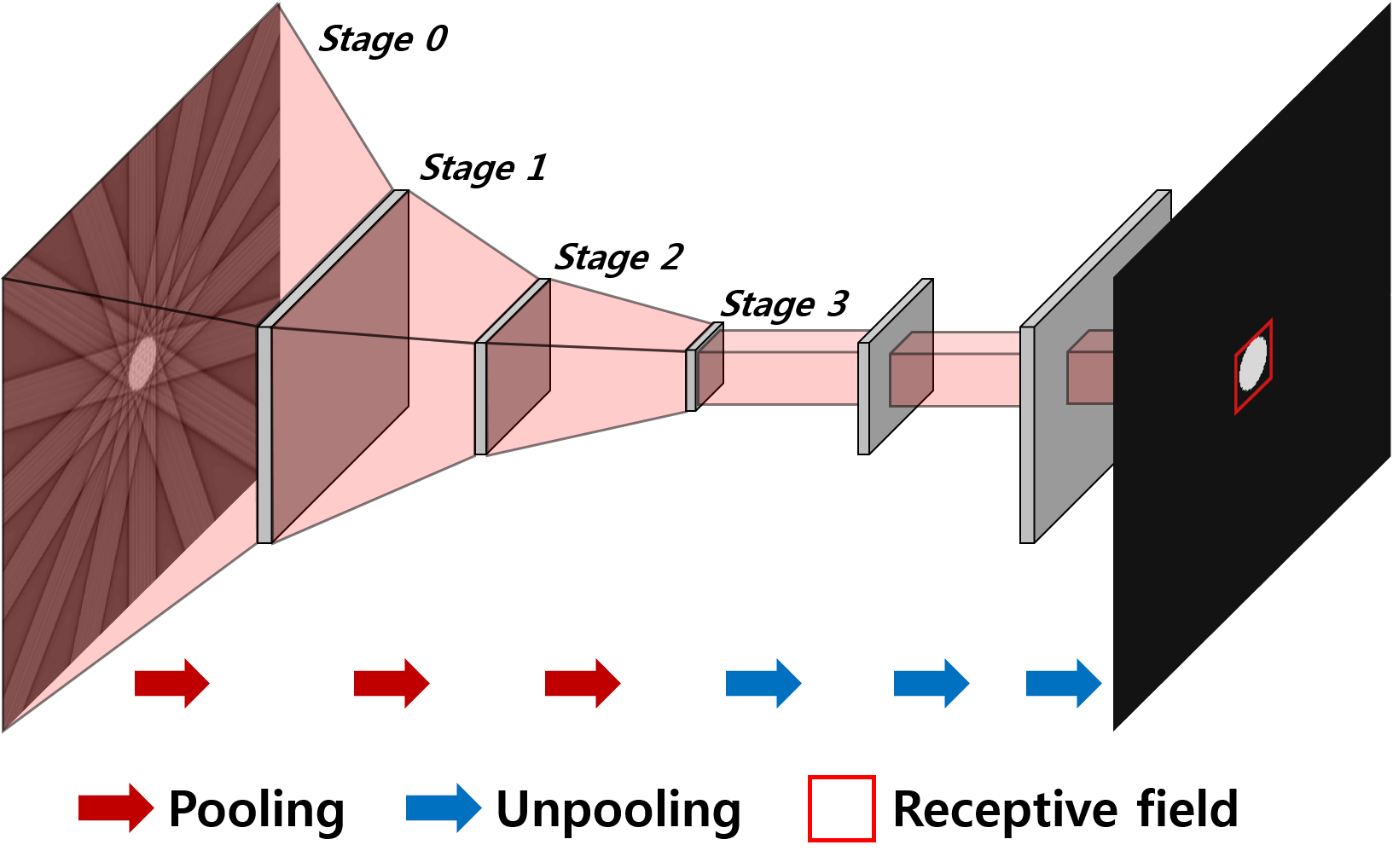}}
    \centerline{\mbox{(a) }\hspace{4.cm}\mbox{(b)}}
            \vspace*{-0.3cm}
    \caption{Effective receptive field comparison. (a) Single resolution CNN without pooling, and (b) U-Net.}
    \label{fig:receptive_field}
\end{figure}

One of the most important characteristics of  multi-resolution architecture like U-Net \cite{ronneberger2015u} is the exponentially large receptive field due to
the pooling and unpooling layers.
For example, 
Fig. \ref{fig:receptive_field} compares  the network depth-wise effective receptive field of a multi-resolution network
and a baseline single resolution network without pooling layers.
With the same size convolutional filters, the effective receptive field is enlarged in the network with pooling layers.
Thus,  the multi-resolution architecture is good for 
the sparse view CT reconstruction to deal with the globally distributed streaking artifacts \cite{jin2016deep,han2016deep}.

\begin{figure}[!b]
   \vspace{-0.5cm}
    \centerline{\includegraphics[width=0.9\linewidth]{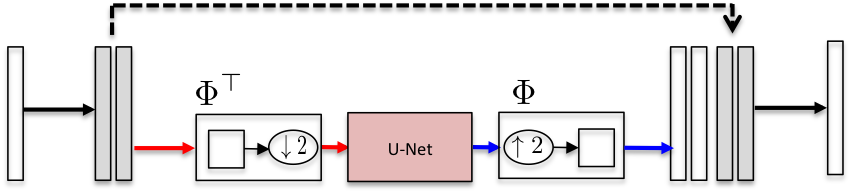}}
    \centerline{\mbox{(a)}}
    \vspace{0.1cm}
        \centerline{\includegraphics[width=0.9\linewidth]{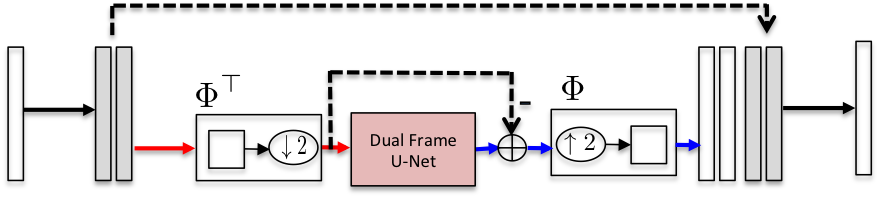}}
    \centerline{\mbox{(b)}}
        \vspace{0.1cm}
            \centerline{\includegraphics[width=0.9\linewidth]{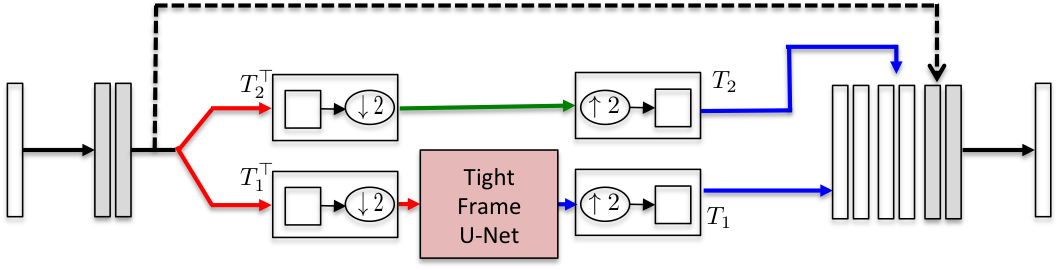}}
    \centerline{\mbox{(c)}}
    \vspace*{-0.2cm}
    \caption{Simplified  U-Net architecture and its variants. (a) Standard U-Net, (b) dual frame U-Net, and (c) tight frame U-Net with concatenation. 
    Dashed lines refer to the skipped-connection, square-box within $\Phi,\Phi^\top$ and $T_k,T_k^\top$ correspond to the sub-band filters. The next level U-Net units are  added recursively to the low-frequency band signals.}
    \label{fig:Unet}
\end{figure}

To understand U-Net in detail, consider a  simplified U-Net architecture  illustrated in
Fig.~\ref{fig:Unet}(a), where the next level  U-Net is recursively applied to the low-resolution signal (for the 2-D implementation,
see Fig.~\ref{fig:architecture}(a)).
Here, the input $f\in \Rd^n$ is first filtered with local convolutional filters $\overline\Psi$, which is then reduced to a half size approximate
signal using a pooling operation $\Phi$.
Mathematically, this step can be represented by
\begin{eqnarray}\label{eq:CUnet}
C = \Phi^\top (f\circledast\overline\Psi) = \Phi^\top \hank_d(f) \Psi \ ,
\end{eqnarray}
where $f\circledast \overline\Psi$ denotes the multi-channel convolution in CNN.
For the case of average pooing,
$\Phi^\top$ denotes a pooling operator given by
\begin{eqnarray}\label{eq:H}
\Phi^\top  =  
 \frac{1}{\sqrt{2}}  \begin{bmatrix} 1 & 1 &  0 & 0 & \cdots & 0  & 0 \\  0 & 0 & 1 & 1 & \cdots & 0  & \\
 & \vdots &   &  & \ddots & \vdots &   \\
  0 & 0 &  0 & 0 & \cdots & 1 & 1 \end{bmatrix} ~ \in \Rd^{\frac{n}{2}\times n} \  .
\end{eqnarray}
The U-Net has the by-pass connection to compensate for the lost high frequency detail during pooling (see Fig.~\ref{fig:Unet}(a) and its 2-D
implementation in Fig.~\ref{fig:architecture}(a)).
Combining the two, the convolutional framelet coefficients can be  represented by
\begin{eqnarray}\label{eq:Y}
C_{ext}  &=& \Phi_{ext}^\top  (f\circledast \Psi)  = \begin{bmatrix} B \\ S  \end{bmatrix} ,
\end{eqnarray}
where  $\Phi_{ext}^\top $ refers to the extended pooling:
\begin{eqnarray}\label{eq:A}
\Phi_{ext}^\top := \begin{bmatrix} I   \\ \Phi^\top \end{bmatrix}, 
\end{eqnarray}
and the bypass component $B$ and the low pass subband $S$ are given by
\begin{eqnarray}
B = f\circledast \overline\Psi, \quad  S = \Phi^\top  (f\circledast \overline\Psi) .
\end{eqnarray}
Accordingly, we have
\begin{eqnarray}
\Phi_{ext}\Phi_{ext}^\top  = I + \Phi \Phi^\top ,
\end{eqnarray}
where  $\Phi\Phi^\top=  P_{R(\Phi)}$ for the case of average pooling. Thus, 
$\Phi_{ext}$ does not satisfy the frame condition \eqref{eq:framecond}, which results in artifacts.
In particular, we have shown in our companion paper \cite{ye2017deep} that this leads to an overemphasis of the low frequency components of images due to the duplication of the low frequency branch. 
See \cite{ye2017deep} for more details.

\subsection{Dual Frame U-Net}

\begin{figure*}[!t]
    \centerline{\includegraphics[width=0.9\linewidth]{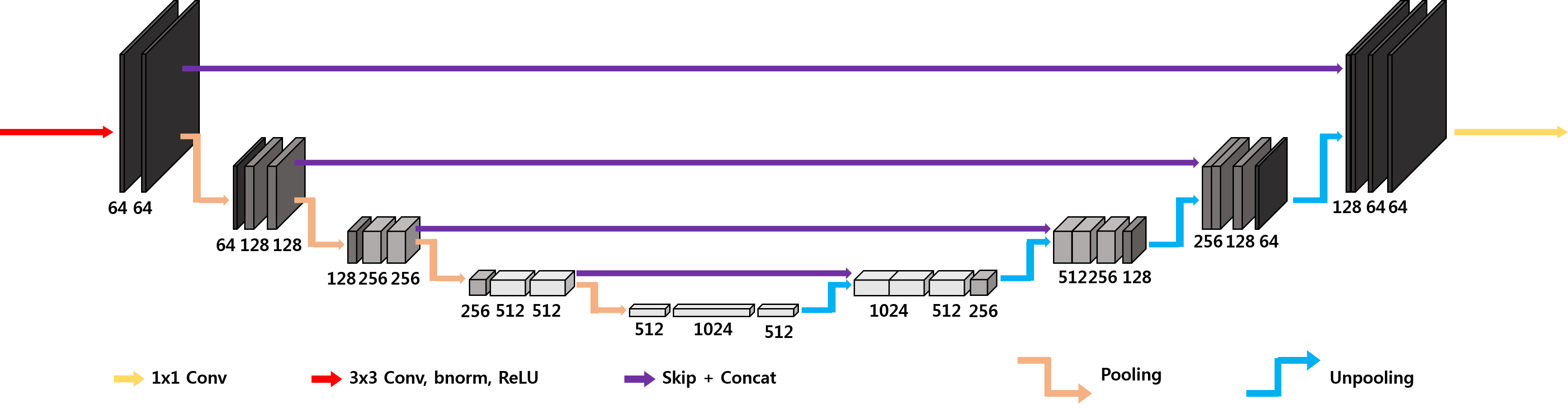}}
    \centerline{\mbox{(a)}}
        \centerline{\includegraphics[width=0.9\linewidth]{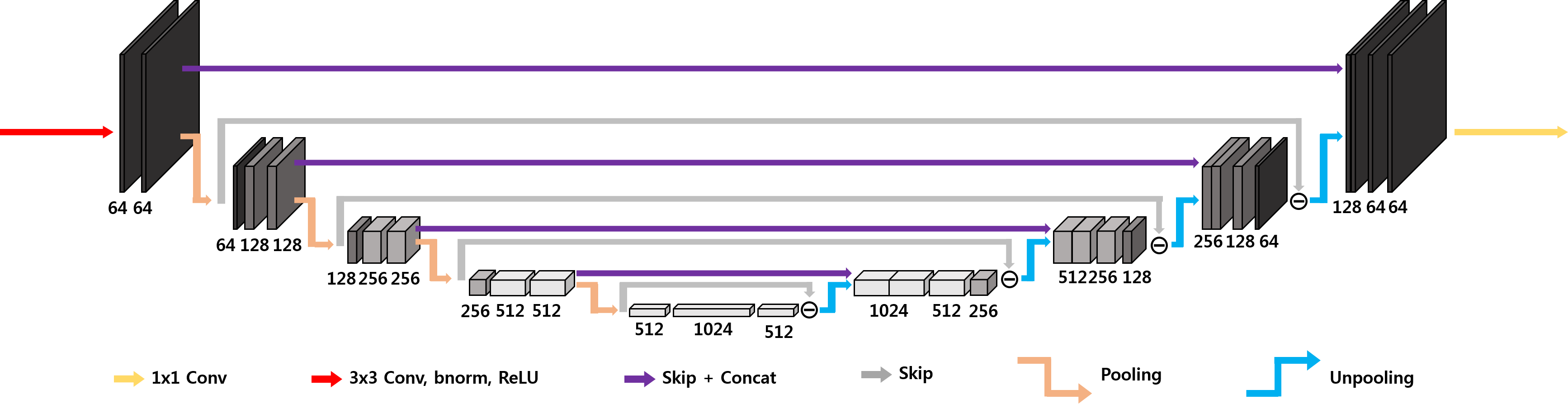}}
    \centerline{\mbox{(b)}}
            \centerline{\includegraphics[width=0.9\linewidth]{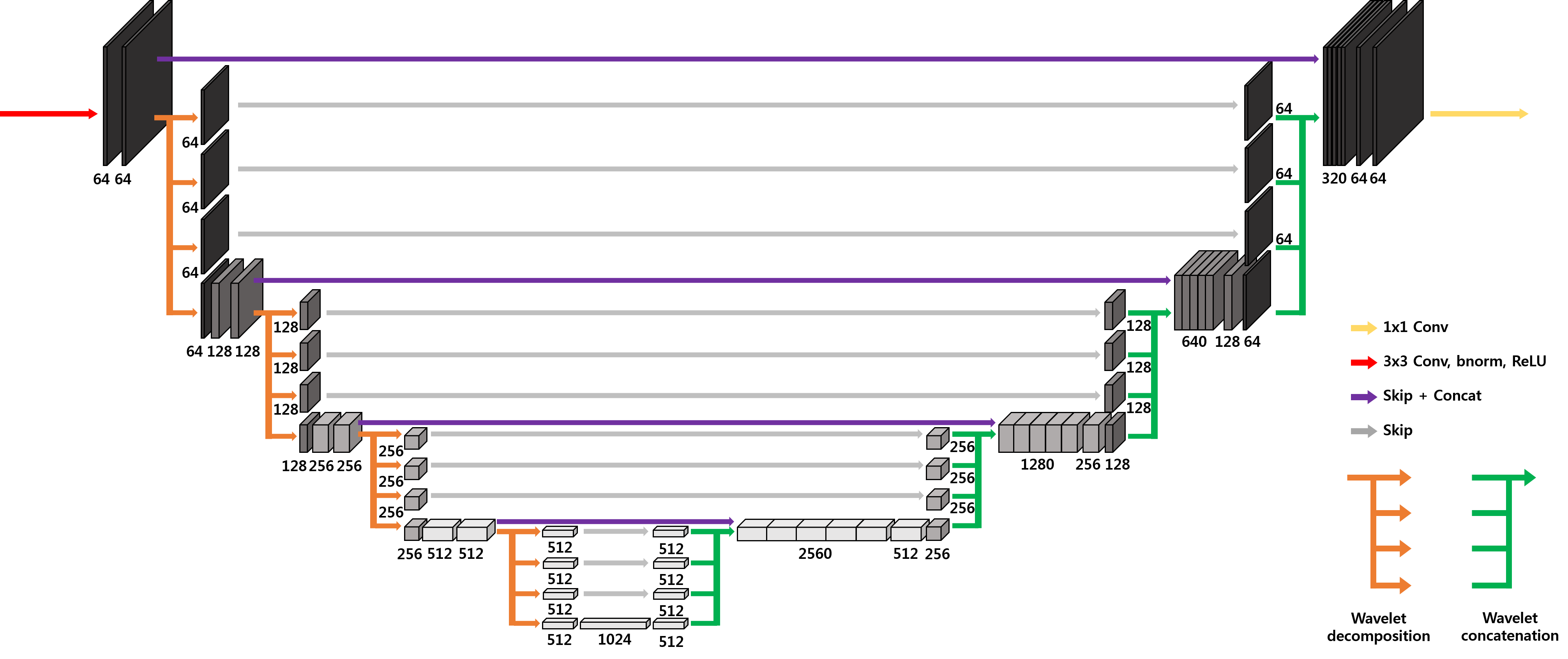}}
    \centerline{\mbox{(c)}}
    \caption{Simplified  U-Net architecture and its variants. (a) Standard U-Net, (b) dual frame U-Net, and (c) tight frame U-Net with concatenation. 
    Dashed lines refer to the skipped-connection, square-box within $\Phi,\Phi^\top$ and $T_k,T_k^\top$ correspond to the sub-band filters. The next level U-Net units are recursively added to the low-frequency band signals.}
    \label{fig:architecture}
\end{figure*}

One simple fix for the aforementioned limitation is using the dual frame. 
Specifically, using \eqref{eq:dual}, the dual frame for $\Phi_{ext}$ in \eqref{eq:A} can be obtained as follows:
\begin{eqnarray}
\tilde \Phi_{ext} =  (\Phi_{ext}\Phi_{ext}^\top)^{-1}\Phi_{ext}=  ( I + \Phi \Phi^\top)^{-1} \begin{bmatrix} I  & \Phi \end{bmatrix} .
\end{eqnarray}
Thanks to the the matrix inversion lemma and the orthogonality $\Phi^\top\Phi =I$ for the case of average pooling, we have
\begin{eqnarray}
 ( I + \Phi \Phi^\top)^{-1} = I - \Phi (I + \Phi^\top \Phi)^{-1} \Phi^\top = I - \frac{1}{2} \Phi \Phi^\top .
\end{eqnarray}
Thus, the dual frame is given by
\begin{eqnarray}
\tilde \Phi_{ext}
= \left(I -\Phi \Phi^\top/2 \right) \begin{bmatrix} I  & \Phi \end{bmatrix} = \begin{bmatrix} I-\Phi\Phi^\top/2 & \Phi/2 \end{bmatrix} .
\end{eqnarray}
For a given framelet coefficients $C_{ext}$ in \eqref{eq:Y},  the reconstruction using the dual frame is then given by
\begin{eqnarray}
\hat C_{ext}  := \tilde \Phi_{ext} C_{ext} 
&= & \left( I-\frac{\Phi\Phi^\top}{2}\right) B + \frac{1}{2} \Phi S  \label{eq:dualUnet}
 \\
&=& B + \frac{1}{2} \underbrace{\Phi}_{\mbox{unpooling}} \overbrace{(S - \Phi^\top B)}^{\mbox{residual}} . \notag
 \end{eqnarray}
Eq.~\eqref{eq:dualUnet} suggests a network structure for the dual frame U-Net.
More specifically,  unlike the U-Net,  the {\em residual signal} at the low resolution is upsampled through the unpooling layer.
This can be easily implemented using additional by-pass connection for the low-resolution signal as shown in 
 Fig.~\ref{fig:Unet}(b) and its 2-D
implementation in Fig.~\ref{fig:architecture}(b).  
This simple fix allows our network to satisfy the frame condition \eqref{eq:framecond}. 
However, there exists  noise amplification   from the
condition number of $I+\Phi\Phi^\top=I+ P_{R(\Phi)}$, which is equal to 2.

Similar to the U-Net, the final step of dual frame U-Net is the concatenation and the multi-channel convolution,  which is
equivalent to applying the inverse Hankel operation, i.e. $\hank_d^\dag(\cdot)$, to
the processed framelet coefficients multiplied with the local basis \cite{ye2017deep}. 
Specifically, the concatenated signal is given by
\begin{eqnarray}
W  &=& \begin{bmatrix} B & \frac{1}{2} {\Phi} {(S - \Phi^\top B)} \end{bmatrix} .
\end{eqnarray}
The final convolution is equivalently computed by
\begin{eqnarray}
\hat f &=& \hank_d^\dag \left(W \begin{bmatrix} \Xi^\top \\ \Theta^\top \end{bmatrix} \right)  \notag \\
&=&\hank_d^\dag (B\Xi^\top) +\frac{1}{2} \hank_d^\dag (\Phi S \Theta^\top)-\frac{1}{2} \hank_d^\dag (\Phi \Phi^\top B \Theta^\top) \notag \\
&=&  \hank_d^\dag (\hank_d(f)\Psi\Xi^\top)  \notag \\
&=& \frac{1}{d} \sum_{i=1}^q \left( f \circledast \overline\psi_i \circledast \xi_i \right) ,\label{eq:fdual}
\end{eqnarray}
where the third equality comes from $S=\Phi^\top (f\circledast \overline\Psi) = \Phi^\top B$.
Therefore, by choosing the  local filter basis such that $\Psi\Xi^\top=I$, the right hand side of \eqref{eq:fdual} becomes equal to $f$,
satisfying
the recovery condition.

\subsection{Tight Frame U-Net}

Another way to improve the performance of U-Net with minimum noise amplification is using tight filter-bank frames or wavelets.
Specifically,   the non-local basis  $\Phi^\top$ is now composed of filter bank:
\begin{eqnarray}
\Phi &=& \begin{bmatrix} T_1  & \cdots & T_L \end{bmatrix},
\end{eqnarray}
where
$T_k$ denotes the   $k$-th subband  operator.  We further assume that the filter bank is tight, i.e.
\begin{eqnarray}\label{eq:tightT}
\Phi\Phi^\top = \sum_{k=1}^L T_k T_k^\top = c I,
\end{eqnarray}
for some scalar $c>0$.
Then, the convolutional framelet coefficients including a by-pass connection can be written by
\begin{eqnarray}
C_{ext} := \Phi_{ext}^\top  (f\circledast \Psi)  =  \begin{bmatrix} B^\top & S_1^\top & \cdots  &S_L^\top \end{bmatrix}^\top ,
\end{eqnarray}
where
\begin{eqnarray}
\Phi_{ext}:= \begin{bmatrix} I  & T_1  & \cdots  &T_L \end{bmatrix}^\top, ~ B=f\circledast \Psi,~S_k = T_k^\top C  \ .
\end{eqnarray}
Now, we can easily see that $\Phi_{ext}$ is also a tight frame, since
\begin{eqnarray}
\Phi_{ext}\Phi_{ext}^\top  = I + \sum_{k=1}^L T_k T_k^\top = (c+1) I \ . 
\end{eqnarray}

There are several important tight filter bank frames.  One of the most simplest one is that Haar wavelet transform with low and high sub-band decomposition,
where $T_1$ is the low-pass subband, which is equivalent to the average pooling in \eqref{eq:H}. Then, $T_2$ is the high pass filtering given by
\begin{eqnarray}\label{eq:T2}
T_2 =  \frac{1}{\sqrt{2}}  \begin{bmatrix} 1 & -1 &  0 & 0 & \cdots & 0  & 0 \\  0 & 0 & 1 & -1 & \cdots & 0  & \\
 & \vdots &   &  & \ddots & \vdots &   \\
  0 & 0 &  0 & 0 & \cdots & 1 & -1 \end{bmatrix}^\top
\end{eqnarray}
and we can easily see that 
$T_1T_1^\top + T_2T_2^\top = I,$
so the Haar wavelet frame is tight.
The corresponding tight frame U-Net structure is illustrated in Fig.~\ref{fig:Unet}(c) and and its 2-D
implementation in Fig.~\ref{fig:architecture}(c).
In contrast to the standard U-Net, there is an additional high-pass branch.  
Similar to the original U-Net,  in our tight frame U-Net,
each subband signal is by-passed to the individual concatenation layers as shown in Fig.~\ref{fig:Unet}(c) and its 2-D
implementation in Fig.~\ref{fig:architecture}(c).  
Then,  the convolutional layer after the concatenation can provide weighted sum whose weights are learned from data.
This simple fix makes the frame tight.

In the following, we examine the performance of U-Net and its variation  for sparse-view CT, where the globally
distributed streaking artifacts require  multi-scale deep networks.

\section{Methods}

\subsection{Data Set}

As a training data, we used  ten patient data  provided by AAPM Low Dose CT Grand Challenge
(http://www.aapm.org/GrandChallenge/LowDoseCT/). 
From the images  reconstructed from  projection data,   720 synthetic projection data were generated by re-projecting using $radon$ operator in MATLAB. Artifact-free original images were reconstructed by $iradon$ operator in MATLAB using all 720 views. Sparse-view input images  were generated using $iradon$ operator from 60, 90,120, 180, 240, and 360 projection views, respectively. These sparse view reconstruction images correspond to each downsampling factor x12, x8, x6, x4, x3, and x2. For our experiments,  the label images were defined as the difference between the sparse view reconstruction and the full view reconstruction.

Among the ten patient data,  eight patient data were used for training and one patient data was for validation, whereas the remaining one was used for test. 
{This corresponds to  3720 slices of $512\times 512$ images for the training data, and 254 slices of  $512\times 512$ images for the validation data. The test data was 486 slices of $512\times 512$ images.}
The training data was augmented by conducting horizontal and vertical flipping.
For the training data set, we used the 2-D FBP reconstruction using  60, 120 and 240 projection views simultaneously as input, and the residual
image between the full view (720 views) reconstruction and the sparse view reconstructions were used as label.
For quantitative evaluation, the normalized mean square error (NMSE) value was used, which is defined as
\begin{eqnarray}
	NMSE = \frac{\sum_{i=1}^{M} \sum_{j=1}^{N} [f^*(i,j) - \hat{f}(i, j)]^2}{\sum_{i=1}^{M}\sum_{j=1}^{N}[f^*(i,j)]^2},
\end{eqnarray}
where $\hat{f}$ and $f^*$ denote the reconstructed images and ground truth, respectively. $M$ and $N$ are the number of pixel for row and column.
We also use the peak signal to noise ratio (PSNR), which is defined by
\begin{eqnarray}
	PSNR 
		 &=& 20 \cdot \log_{10} \left(\frac{NM\|f^*\|_\infty}{\|\hat f- f^*\|_2}\right) \  .
\label{eq:psnr}		 
\end{eqnarray}
We also used the structural similarity (SSIM) index  \cite{wang2004image}, defined as
\begin{equation}
	SSIM = \dfrac{(2\mu_{\hat f}\mu_{f^*}+c_1)(2\sigma_{\hat f f^*}+c_2)}{(\mu_{\hat f}^2+\mu_{f^*}^2+c_1)(\sigma_{\hat f}^2+\sigma_{f^*}^2+c_2)},
\end{equation}
where $\mu_{\hat f}$ is a average of $\hat f$, $\sigma_{\hat f}^2$ is a variance of $\hat f$ and $\sigma_{\hat f f^*}$ is a covariance of $\hat f$ and $f^*$. 
There are two variables to stabilize the division such as $c_1=(k_1L)^2$ and $c_2=(k_2L)^2$.
$L$ is a dynamic range of the pixel intensities. $k_1$ and $k_2$ are constants by default $k_1=0.01$ and $k_2=0.03$.

\subsection{Network Architecture}

As shown in Figs. \ref{fig:architecture}(a)(b)(c),
the original,  dual frame and tight frame U-Nets  consist of convolution layer, batch normalization~\cite{ioffe2015batch}, rectified linear unit (ReLU) \cite{krizhevsky2012imagenet}, and contracting path connection with concatenation~\cite{ronneberger2015u}.
Specifically,
 each stage contains  four sequential layers composed of convolution with $3\times3$ kernels, batch normalization, and ReLU layers.
 Finally,  the last stage has two sequential layers and the last layer contains only convolution layer with $1\times1$ kernel. 
The number of channels for each convolution layer is illustrated in  Figs. \ref{fig:architecture}(a)(b)(c).
Note that the number of channels are doubled after each pooling layers.
The differences between the original, dual frame and the tight frame U-Net are from the pooling and unpooling layers. 

\begin{figure*}[!hbt]
    \centerline{\includegraphics[width=0.90\linewidth]{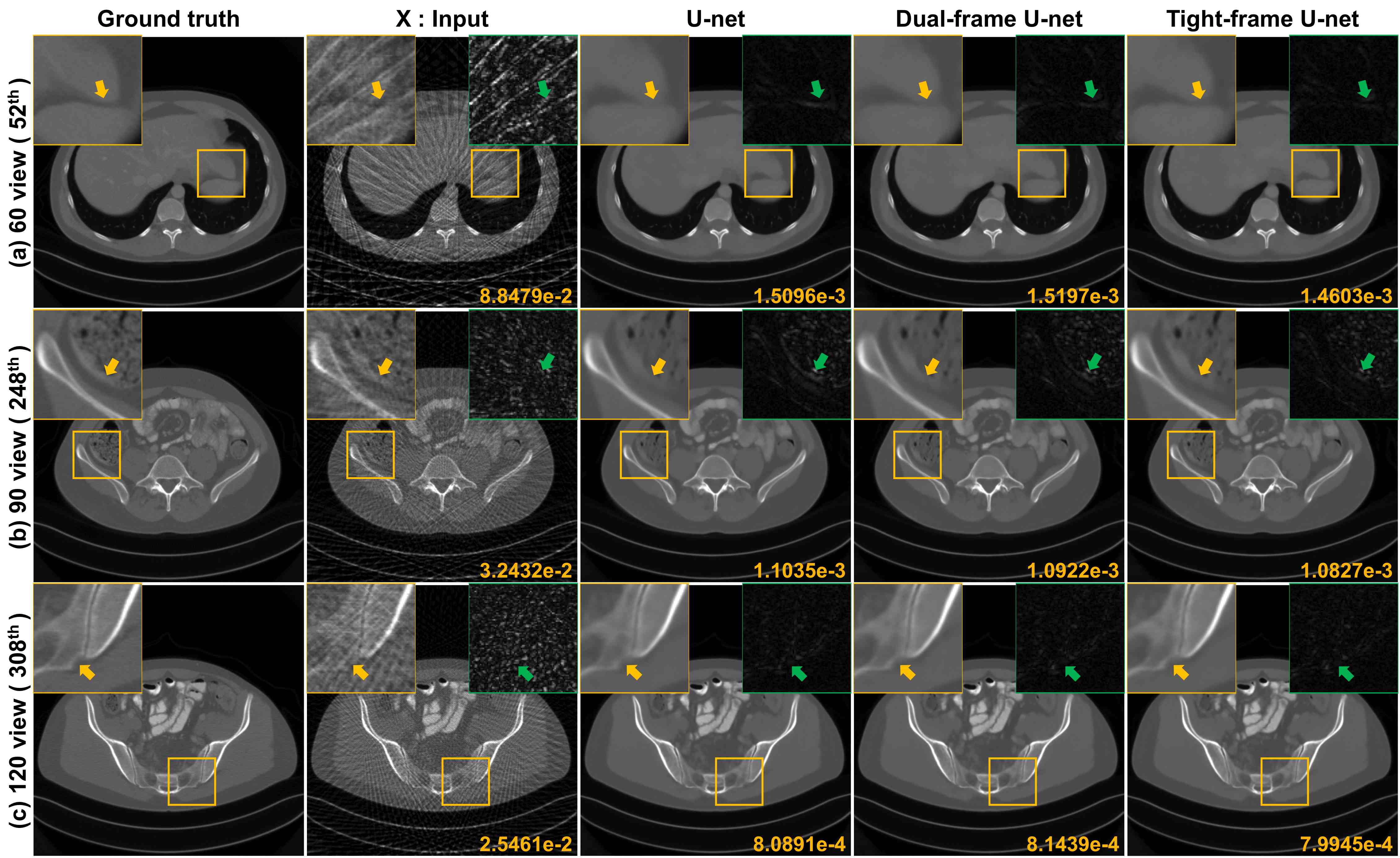}}
    \caption{Reconstruction results by original, dual frame and tight frame U-Nets at various sparse view reconstruction. Yellow and green boxes illustrate the enlarged view and the difference images, respectively. The number written to the images is the NMSE value.}
    \label{fig:frame_result}
\end{figure*}


\subsection{Network  training}
The proposed network was trained by stochastic gradient descent (SGD).  The regularization parameter was $\lambda = 10^{-4}$. The learning rate was set from $10^{-3}$ to $10^{-5}$ which was gradually reduced at each epoch. The number of epoch was 150. A mini-batch data using image patch was used,  and the size of image patch was $256\times256$. 
Since the convolution filters are spatially invariant, we can
use these filters in the inferencing stage. In this case,  the input size is $512 \times 512$.

 
The network was implemented using MatConvNet toolbox (ver.24) \cite{vedaldi2015matconvnet} in MATLAB 2015a environment (Mathwork, Natick). We used a GTX 1080 Ti graphic processor and i7-7700 CPU (3.60GHz). The network takes about 4 day for training.

\begin{table}[h!] 
\caption{Quantitative comparison of different  methods.}
\vspace*{-0.5cm}
\label{tbl:NMSE}
\begin{center}
\begin{adjustbox}{width=0.48\textwidth}
{\begin{tabular}{c|cccccc}
\hline
PSNR [dB]	& 60 views & 90 views & 120 views &  180 views & 240 views & 360 views\\
		(whole image area)												& ( x12 )	& ( x8 )	& ( x6 )	& ( x4 )	& ( x3 )	& ( x2 )	\\
\hline
FBP													& 22.2787	& 25.3070 & 27.4840 & 31.8291 & 35.0178 & 40.6892 \\
U-Net												& 38.8122	& 40.4124 & 41.9699 & 43.0939 & 44.3413 & {45.2366} \\
Dual frame U-Net  					& 38.7871	& 40.4021 & 41.9397 & 43.0795 & 44.3211 &  \textbf{45.2816} \\
Tight frame U-Net						& \textbf{38.9218}	& \textbf{40.5091} & \textbf{42.0457} & \textbf{43.1800} & \textbf{44.3952} & 
{45.2552} \\
\hline
\hline
PSNR [dB]	& 60 views & 90 views & 120 views &  180 views & 240 views & 360 views\\
			(within body)											& ( x12 )	& ( x8 )	& ( x6 )	& ( x4 )	& ( x3 ) & ( x2 )	\\
\hline
FBP													& 28.9182	& 32.0717 & 33.8028 &  38.2559 & 40.7448 & 45.4611 \\
U-Net												& 40.3733	& 42.1512 & 43.6840 & 44.9418 & 46.4402 & 47.5937 \\
Dual frame U-Net
    & 40.3775	& 42.1462 & 43.6973 & 44.9717 & 46.4653 & \textbf{47.6765} \\
Tight frame U-Net						
    & \textbf{40.4856}	& \textbf{42.2380} & \textbf{43.7682} & \textbf{45.0406} & \textbf{46.4847} & 47.5797 \\
\hline

\end{tabular}}
\end{adjustbox}
\end{center}
\end{table}

\section{Experimental Results}

\begin{figure*}[!hbt]
    \centerline{\includegraphics[width=0.90\linewidth]{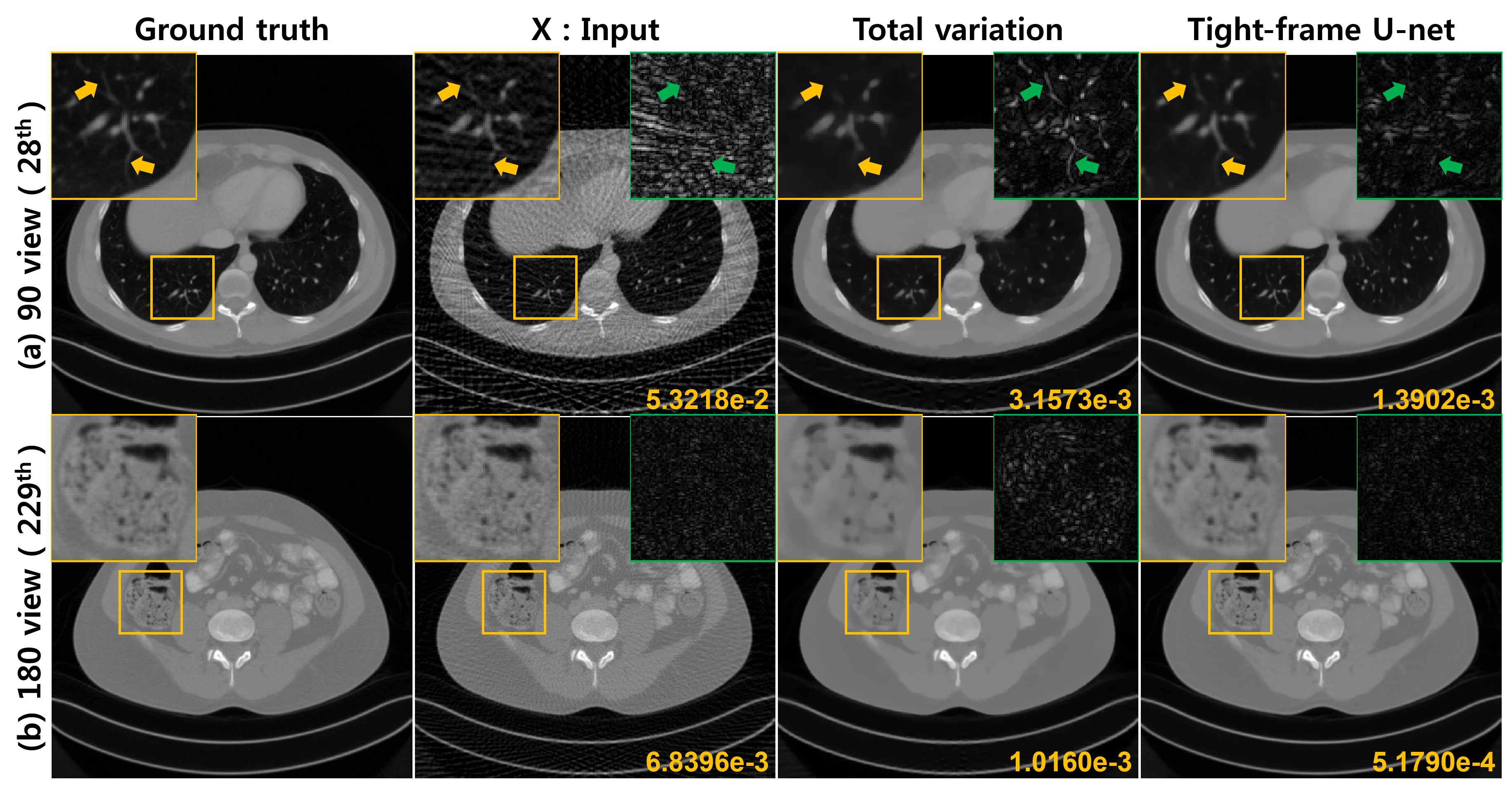}}
    \caption{Reconstruction follows TV method and the proposed tight frame U-Net. Yellow and green boxes illustrate the enlarged view and the difference images, respectively. The number written to the images is the NMSE value.}\label{fig:proposed_result}
\end{figure*}

In Table~\ref{tbl:NMSE}, we give the average PSNR values of U-Net and its variants when applied to sparse view CT from different projection views.
All methods offer significant gain over the FBP.
Among the three types of U-Net variants, the tight frame U-Net produced the best PSNR values, followed by the standard U-Net.
However, if we restrict the ROI within the body area by removing the  background and patient bed, 
the tight frame U-Net was best, which is followed by the dual frame U-Net.
It is also interesting to see that the dual frame U-Net was the best for the x2 downsampling factor.
This implies that the proposed U-Net variants provide quantitatively better reconstruction quality over the standard U-Net.

In addition,  the visual inspection provides advantages of  our U-Net variants.
Specifically,
Fig.~\ref{fig:frame_result} compares the reconstruction results by original, dual frame, and tight frame U-Nets.  
As shown in the enlarged images and the difference images, the U-Net produces blurred edge images in many areas, while the dual frame and tight frame U-Nets enhance the high frequency characteristics of the images.
Despite the better subjective quality, the reason that 
 dual frame U-Net  in the case of whole image area does not offer better PSNR values than the standard U-Net in Table~\ref{tbl:NMSE} 
 may be due to  the greater noise amplification factor so that the error in  background and patient bed may dominate.
 Moreover, the low-frequency duplication in the standard U-Net
 may contribute the better PSNR values  in this case.
 However,  our tight frame U-Net   not only provides better average PSNR values (see Table~\ref{tbl:NMSE})
 and  the minimum  NMSE values  (see Fig.~\ref{fig:frame_result}), but also
improved visual quality  over the standard U-Net.
Thus, we use the tight frame U-Net in all other experiments.


\begin{figure}[!b]
    \centerline{\includegraphics[width=0.9\linewidth,height=11cm]{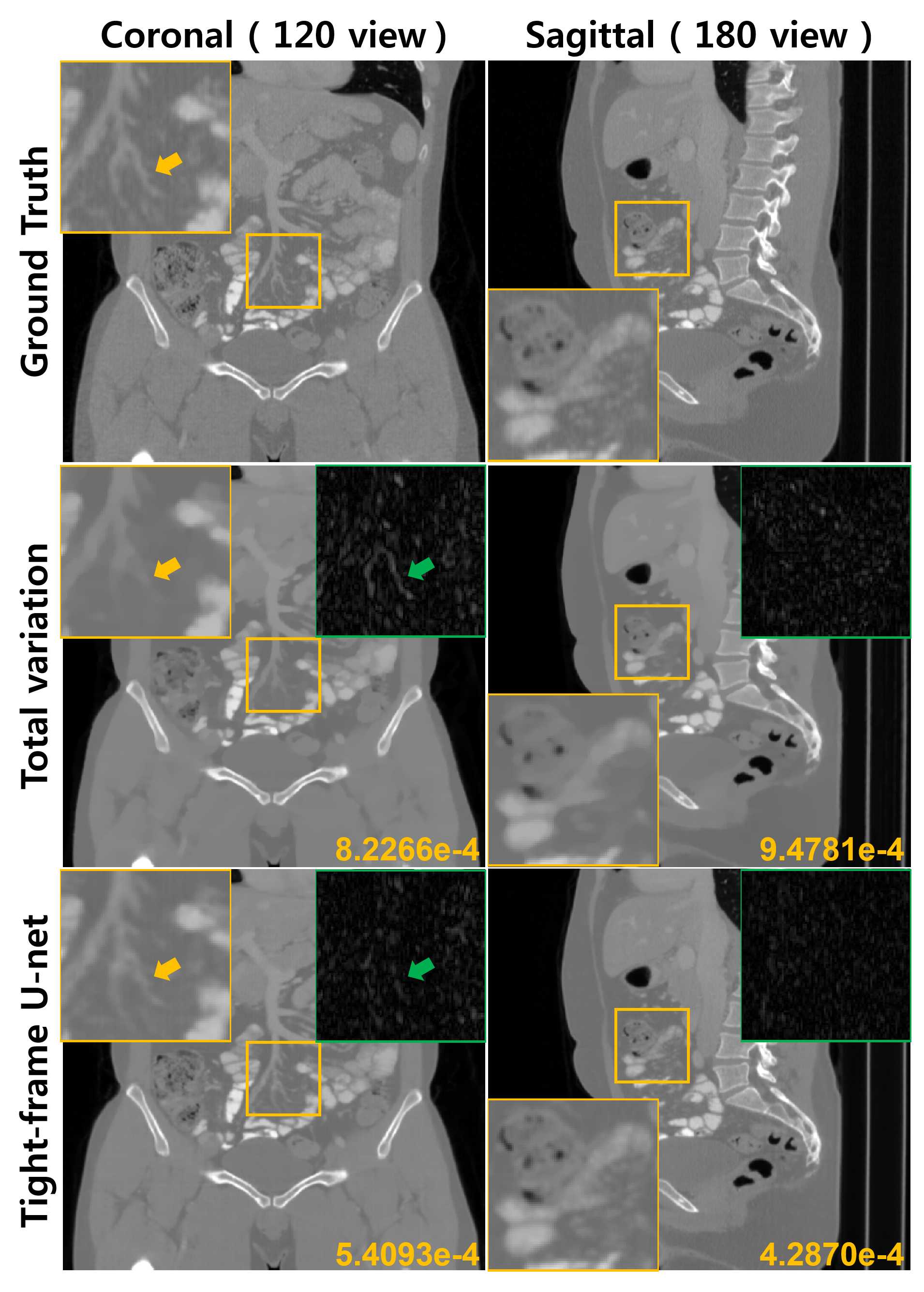}}
    \caption{Coronal and sagittal views of the reconstruction method according to the TV method and the proposed tight frame U-Net. Yellow and green boxes illustrate the enlarged viewand  difference pictures. The number written to the images is the NMSE value.}
    \label{fig:cutview_result}
\end{figure}

\begin{figure}[!t]
    \centerline{\includegraphics[width=1\linewidth]{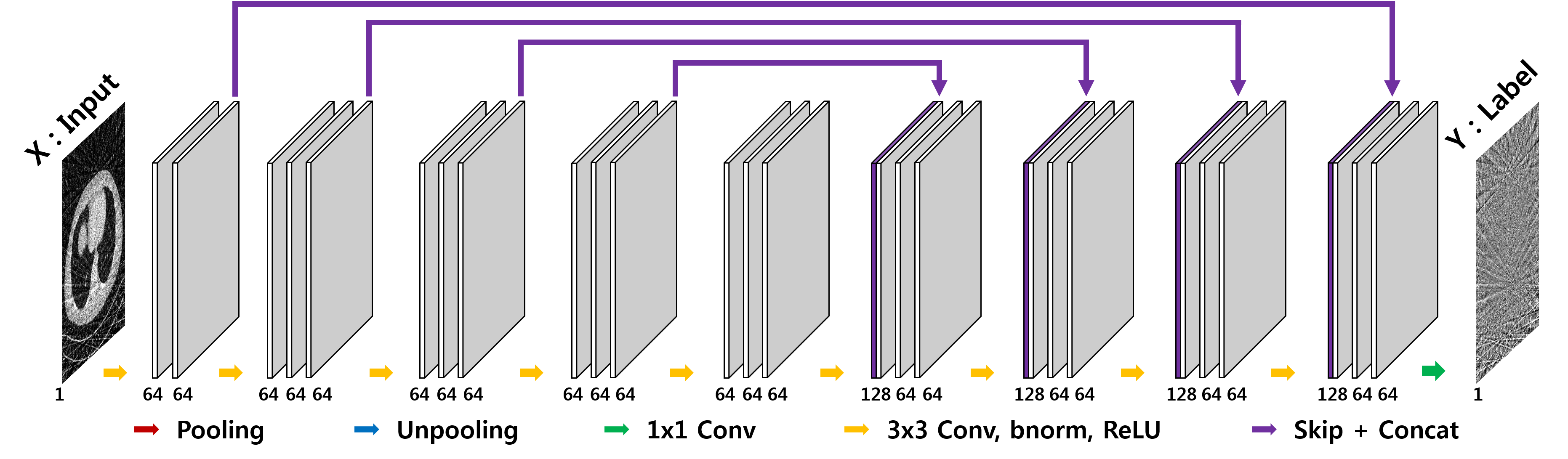}}
    \caption{Single scale baseline network.}
    \label{fig:compared_network}
\end{figure}

Figs. \ref{fig:proposed_result}(a)(b) compared the reconstruction results by the proposed method and TV from 90 and 180 projection views, respectively. 
The TV method is formulated as follows:
\begin{eqnarray}
	\arg \min_{x} \frac{1}{2} || y - Af ||_{2}^{2} + \lambda TV(f),
\end{eqnarray}
where $f$ and $y$ denote the reconstructed images and the measured sinogram and $A$ is projection matrix.
The regularization parameter $\lambda$  was chosen by trial and error to get the best trade-off between the resolution and NMSE values, resulting in 
a value of $5\times10^{-3}$. The TV method was solved by Alternating Direction Method of Multipliers (ADMM) optimizer \cite{ramani2012splitting}. As the number of projection views decreases, we have observed that the number of iterations should gradually increase; 60, 120, and 240 for the algorithm to converge when the number of views is 180, 120, and 90, respectively.

The results  in Fig. \ref{fig:proposed_result}(a)(b) clearly showed that the proposed network removes most of streaking artifact patterns and  preserves detailed structures of underlying images. 
The magnified and difference views in Fig. \ref{fig:proposed_result}(a)(b) confirmed that the detailed structures are very well reconstructed using the proposed method.
On the other hand,  TV method does not provide accurate reconstruction.
Fig. \ref{fig:cutview_result} shows reconstruction results  from coronal and sagittal directions. Accurate reconstruction were obtained using the proposed method.
Moreover, compared to the TV method, the proposed results in Fig.~\ref{fig:proposed_result} and Fig.~\ref{fig:cutview_result}  
provides significantly improved image reconstruction results and much smaller NMSE values.
The average PSNR and SSIM values  in Table~\ref{tbl:TV}  also confirm that the proposed tight frame U-Net consistently outperforms the TV method at all view down-sampling factors.

On the other hand, the computational time for the proposed method is 250 ms/slice with GPU  and 
 5 sec/slice with CPU, respectively, while the
 TV approach in CPU took about 20 $\sim$ 50 sec/slice for reconstruction.
This implies that the proposed method is 4 $\sim$ 10 times faster than the TV approach with significantly better reconstruction performance.

\begin{figure*}[!hbt]
    \centerline{\includegraphics[width=0.90\linewidth]{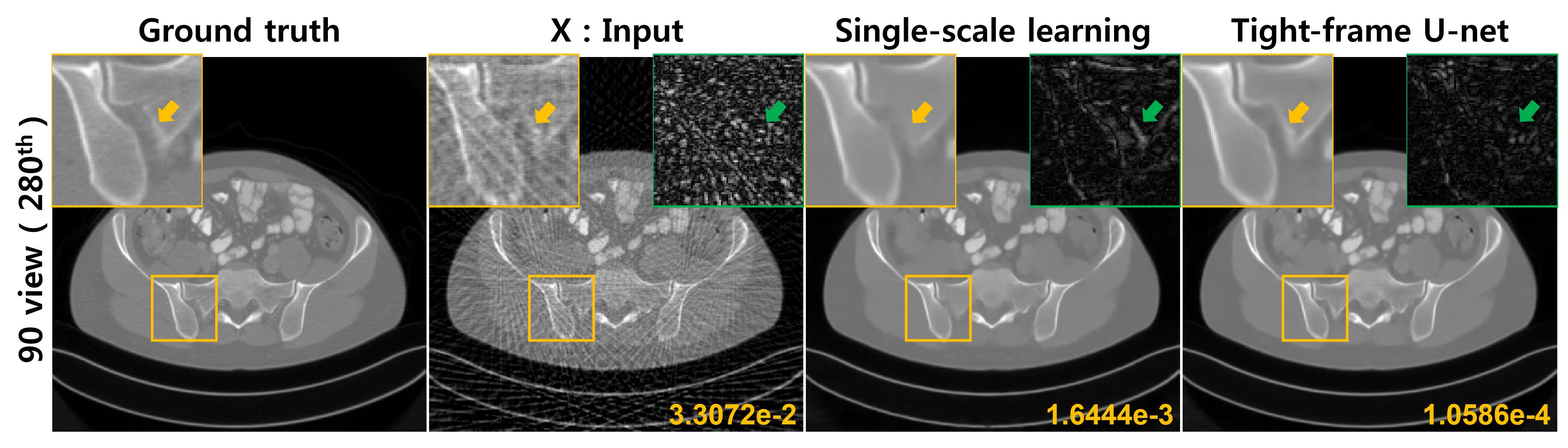}}
            \vspace*{-0.3cm}
    \caption{Reconstruction follows single-scale network and the proposed tight frame U-Net. Yellow and green boxes illustrate the enlarged view and the difference images, respectively. The number written to the images is the NMSE value.}
    \label{fig:single_scale_result}
\end{figure*}

\begin{table}[h!] 
\caption{Quantitative comparison with TV approach.}
\vspace*{-0.5cm}
\label{tbl:TV}
\begin{center}
\begin{adjustbox}{width=0.48\textwidth}
{\begin{tabular}{c|cccccc}
\hline
\multirow{ 2}{*} {PSNR [dB]}	& 60 views & 90 views & 120 views &  180 views & 240 views & 360 views\\
														& ( x12 )	& ( x8 )	& ( x6 )	& ( x4 )	& ( x3 )	& ( x2 )	\\
\hline
TV													& 33.7113 & 37.2407 & 38.4265 & 40.3774 & 41.6626 & 44.2509 \\
Tight frame U-Net						& \textbf{38.9218}	& \textbf{40.5091} & \textbf{42.0457} & \textbf{43.1800} & \textbf{44.3952} & 
\textbf{45.2552} \\
\hline
\hline
\multirow{ 2}{*} {SSIM}	& 60 views & 90 views & 120 views &  180 views & 240 views & 360 views\\
														& ( x12 )	& ( x8 )	& ( x6 )	& ( x4 )	& ( x3 ) & ( x2 )	\\
\hline
TV													& 0.8808 & 0.9186 & 0.9271 & 0.9405 & 0.9476 & 0.9622 \\
Tight frame U-Net						
    & \textbf{0.9276}	& \textbf{0.9434} & \textbf{0.9547} & \textbf{0.9610} & \textbf{0.9678} & \textbf{0.9708} \\
\hline
\end{tabular}}
\end{adjustbox}
\end{center}
\end{table}

\section{Discussion}

\subsection{Single-scale vs. Multi-scale residual learning}

\begin{table}[!b] 
\caption{Quantitative comparison with a single-scale network.}
\vspace*{-0.5cm}
\label{tbl:single}
\begin{center}
\begin{adjustbox}{width=0.48\textwidth}
{\begin{tabular}{c|cccccc}
\hline
\multirow{ 2}{*} {PSNR [dB]}	& 60 views & 90 views & 120 views &  180 views & 240 views & 360 views\\
														& ( x12 )	& ( x8 )	& ( x6 )	& ( x4 )	& ( x3 )	& ( x2 )	\\
\hline
Single-scale CNN													& 36.7422	& 38.5736 & 40.8814 & 42.1607 & 43.7930 & 44.8450 \\
Tight frame U-Net						& \textbf{38.9218}	& \textbf{40.5091} & \textbf{42.0457} & \textbf{43.1800} & \textbf{44.3952} & 
\textbf{45.2552} \\
\hline
\hline
\multirow{ 2}{*} {SSIM}	& 60 views & 90 views & 120 views &  180 views & 240 views & 360 views\\
														& ( x12 )	& ( x8 )	& ( x6 )	& ( x4 )	& ( x3 )	& ( x2 )	\\
\hline
Single-scale CNN										& 0.8728	& 0.9046 & 0.9331 & 0.9453 & 0.9568 & 0.9630 \\
Tight frame U-Net						
    & \textbf{0.9276}	& \textbf{0.9434} & \textbf{0.9547} & \textbf{0.9610} & \textbf{0.9678} & \textbf{0.9708} \\
\hline
\end{tabular}}
\end{adjustbox}
\end{center}
\end{table}

Next,  we investigated the importance of the multi-scale network.  
As a baseline network, a single-scale residual learning network without pooling and unpooling layers as shown in Fig.~\ref{fig:compared_network} was used.
Similar to the proposed method,
the streaking artifact  images were used as the labels. 
For fair comparison, we set the number of network parameters similar to the proposed method by fixing the number of channels at each layer across all the stages.  
%
In Fig.~\ref{fig:single_scale_result},  the image reconstruction quality and the NMSE values provided by the tight frame U-Net  was much improved compared to the 
single resolution network.
The average PSNR and SSIM values in Table~\ref{tbl:single}  show that single scale network is consistently inferior to the tight frame U-Net for all view down-sampling factors.
This is due to the smaller receptive field in a single resolution network, which is difficult to correct globally distributed streaking artifacts.

\subsection{Diversity of training set}

\begin{figure}[!h]
    \centerline{\includegraphics[width=0.90\linewidth]{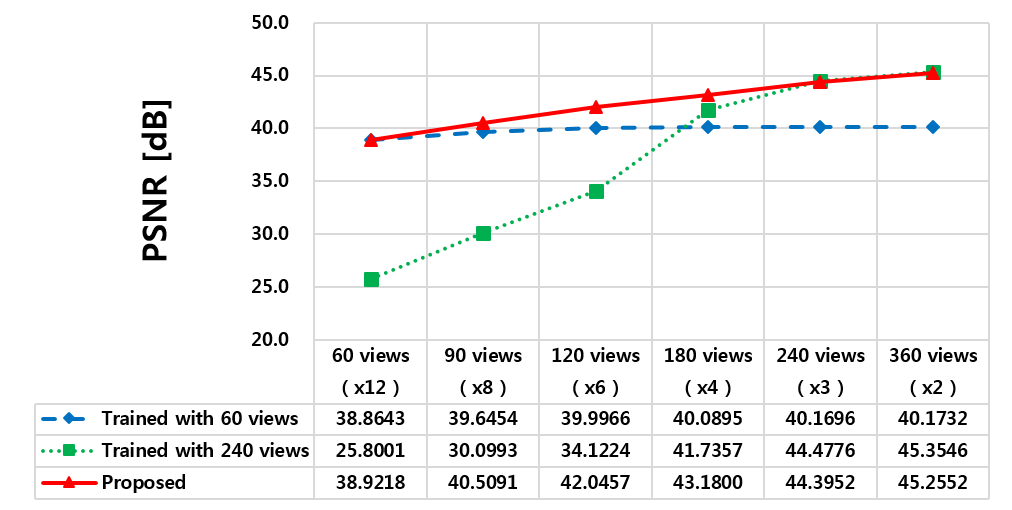}}
        \vspace*{-0.3cm}
    \caption{Quantitative comparison for reconstruction results from the various training set configuration.}
    \label{fig:grp_trained}
\end{figure}

Fig. \ref{fig:grp_trained} shows that average PSNR values of the tight frame U-Net for various view downsampling factors.
Here,  we compared the three distinct training strategies.  First, the tight frame U-Net was trained with the FBP reconstruction
using 60 projection views. The second network was trained using FBP reconstruction from 240 views.
Our proposed network was trained using the FBP reconstruction from 60, 120, and 240 views. 
As shown in  Fig. \ref{fig:grp_trained}, the first two networks provide the competitive performance at 60 and 240 projection views, respectively.
However,  %
 the combined training  offered the best reconstruction across wide ranges of view down-sampling.
 Therefore, to  make the network suitable for  all down-sampling factors,
 we   trained the network  by  using FBP data  from 60, 120, and 240 projection views simultaneously. 
 
%

%

\begin{figure*}[!t]
    \centerline{\includegraphics[width=0.90\linewidth]{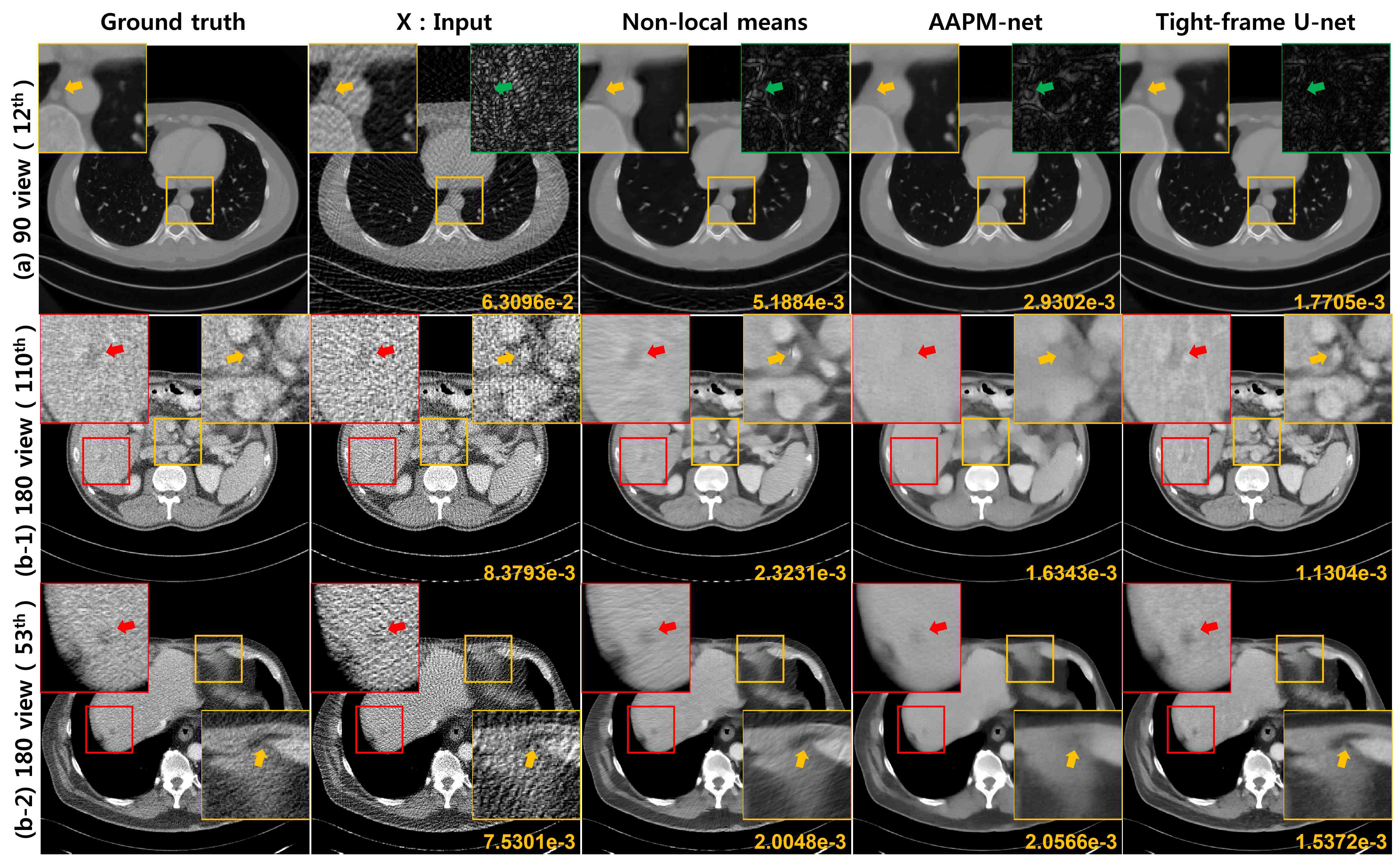}}
    \vspace*{-0.3cm}
    \caption{Reconstruction results by the non-local means \cite{kim2017low}, AAPM-net \cite{kang2016deep} and proposed tight frame U-Net.  (a) 90 view full-dose data, and (b)(c) 180 view quarter-dose data. Yellow and green boxes illustrate the enlarged view and the difference images, respectively. Red boxes indicate the lesion region. The number written to the images is the NMSE value.}
    \label{fig:lesion_case_result}
\end{figure*}

\subsection{Comparison to AAPM Challenge winning algorithms}
Originally, the AAPM low-dose CT Challenge dataset were collected to detect lesions in the quarter-dose CT images, and the dataset consists of full- and quarter-dose CT images. In the Challenge, 
penalized least squares with non-local means penalty \cite{kim2017low} and AAPM-Net \cite{kang2016deep}  were the winners of the  first and the second place,  respectively. 
However, the task in AAPM challenge was to reduce the noises from the tube-current modulated low-dose CT rather than the sparse-view CT.
To demonstrate that a dedicated network is necessary for the sparse-view CT,
we conducted the comparative study for the sparse-view CT using the two winning algorithms at the AAPM challenge.
For a fair comparison, we re-trained the AAPM-Net with the sparse-view CT data,  and the optimal hyper-parameters  for the penalized least squares with non-local means penalty \cite{kim2017low} were determined by trial and error.
Fig. \ref{fig:lesion_case_result}(a) shows that reconstructed images by non-local means, AAPM-Net, and the proposed tight frame U-Net from 90 view full-dose input images. Since the non-local
means algorithm
\cite{kim2017low} and AAPM-Net \cite{kang2016deep} have been designed to  remove noises from tube-current modulated low-dose CT,
their applications results in blurring artifacts. 
The average PSNR and SSIM values in Table~\ref{tbl:aapm}  for 90 view full-dose images confirm that the proposed tight frame U-Net outperforms the AAPM challenge winning algorithms.

We also investigated the lesion detection capability of these algorithms.
In the AAPM challenge,  only quarter-dose images have lesions. Therefore, we generated projection data from the quarter-dose images,
and each algorithm was tested for removing streaking artifacts from 180 view projection data.
As shown in Figs. \ref{fig:lesion_case_result}(b)(c), the non-local means algorithm \cite{kim2017low} and AAPM-Net \cite{kang2016deep} were not good in detecting
the lesions from the streaking artifacts, whereas  the lesion region was clearly detected using the proposed method.
As a byproduct,  the proposed tight frame U-Net successfully removes the low-dose CT noise and offers clear images.

\begin{table}[!h] 
\caption{Quantitative comparison with AAPM Challenge winning algorithms for 90 view reconstruction.}
\vspace*{-0.5cm}
\label{tbl:aapm}
\begin{center}
\begin{adjustbox}{width=0.4\textwidth}
{\begin{tabular}{c|cccccc}
\hline
Algorithm	& Non-local means & AAPM-Net & Tight frame U-Net\\
\hline
PSNR [dB]													& 34.0346	& 38.3493 & \textbf{40.5091} \\
\hline
\hline
Algorithm       	& Non-local means & AAPM-Net & Tight frame U-Net\\
\hline
SSIM 													& 0.8389	& 0.8872 & \textbf{0.9434} \\
\hline
\end{tabular}}
\end{adjustbox}
\end{center}
\end{table}

 \subsection{Max Pooling}
 
In our analysis of U-Net, we consider the average pooling as shown in \eqref{eq:H}, but
we could also define $\Phi^\top$ for the case of the max pooling. In this case, \eqref{eq:H} should be changed as
\begin{eqnarray}\label{eq:Hmax}
 \begin{bmatrix} b_{1,2} & 1-b_{1,2} &  0 & 0 & \cdots & 0  & 0 \\ 
 & \vdots &   &  & \ddots & \vdots &   \\
  0 & 0 &  0 & 0 & \cdots & b_{n-1,n} & 1-b_{n-1,n} \end{bmatrix} ,
\end{eqnarray}
where
\begin{eqnarray}
b_{i,i+1} = \begin{cases} 1, & \mbox{when}~ f[i]=\max\{f[i],f[i+1]\} \\ 0,  & \mbox{otherwise}
\end{cases}.
\end{eqnarray}
To satisfy the frame condition \eqref{eq:framecond},
the corresponding high-pass branch pooling $T_2$ in \eqref{eq:T2} should be
changed accordingly as
\begin{eqnarray}
\begin{bmatrix} 1-b_{1,2} & b_{1,2} &  0 & 0 & \cdots & 0  & 0 \\ 
 & \vdots &   &  & \ddots & \vdots &   \\
  0 & 0 &  0 & 0 & \cdots & 1-b_{n-1,n} & b_{n-1,n} \end{bmatrix} .
\end{eqnarray}
However, we should keep track of all $b_{i,i+1}$ at each step of the pooling, which requires additional memory.
Thus, we are mainly interested in using \eqref{eq:H} and \eqref{eq:T2}.

\section{Conclusion}

In this paper, we  showed that large receptive field network architecture from multi-scale network is essential for sparse view CT reconstruction 
due to the globally distributed streaking artifacts.
Based on the recent theory of deep convolutional framelets, we then showed that the existing U-Net architecture does not meet the frame condition. The resulting disadvantage is often found as the blurry and false image artifacts.
To overcome the limitations,  we proposed dual frame U-Net and tight frame U-Net. While the dual frame U-Net was designed to meet the frame condition, the resulting modification was an intuitive extra skipped connection.
For tight frame U-Net with wavelets, an additional path is needed to process the subband signals. These extra path allows for improved noise robustness
and directional information process, which can be adapted to image statistics. 
Using extensive experiments, we showed that the proposed U-Net variants were better than the conventional U-Net for sparse view CT reconstruction.

\section{Acknowledgment}

The authors would like to thanks Dr. Cynthia McCollough,  the Mayo Clinic, the American Association of Physicists in Medicine (AAPM), and grant EB01705 and EB01785 from the National
Institute of Biomedical Imaging and Bioengineering for providing the Low-Dose CT Grand Challenge data set.
This work is supported by Korea Science and Engineering Foundation, Grant number
NRF-2016R1A2B3008104.
The authors would like to thank Dr. Kyungsang Kim at MGH for providing the code in \cite{kim2017low}.

%


%
%


\end{document}